\documentclass[sigconf]{acmart}
\usepackage{booktabs}
\usepackage{multirow}
\usepackage{graphicx}
\usepackage{subcaption}
\usepackage{enumitem}
\usepackage{colortbl}

\AtBeginDocument{%
  \providecommand\BibTeX{{%
    \normalfont B\kern-0.5em{\scshape i\kern-0.25em b}\kern-0.8em\TeX}}}

\copyrightyear{2024}
\acmYear{2024}
\setcopyright{acmlicensed}\acmConference[KDD '24]{Proceedings of the 30th
ACM SIGKDD Conference on Knowledge Discovery and Data Mining}{August
25--29, 2024}{Barcelona, Spain}
\acmBooktitle{Proceedings of the 30th ACM SIGKDD Conference on Knowledge
Discovery and Data Mining (KDD '24), August 25--29, 2024, Barcelona, Spain}
\acmDOI{10.1145/3637528.3671790}
\acmISBN{979-8-4007-0490-1/24/08}

\begin{document}

\title{Temporal Prototype-Aware Learning for Active Voltage Control on Power Distribution Networks}


\author{Feiyang Xu}
\authornote{Authors contributed equally to this research. E-mail: xufeiyang@zju.edu.cn}
\affiliation{%
  \institution{Polytechnic Institute, Zhejiang University}
  \institution{State Key Laboratory of Blockchain and Security, Zhejiang University}
    \institution{Hangzhou High-Tech Zone (Binjiang) Institute of Blockchain and Data Security}
  \city{Hangzhou}
  \country{China}
  \postcode{310027}
}

\author{Shunyu Liu}
\authornotemark[1]
\authornote{Corresponding author. E-mail: liushunyu@zju.edu.cn}
\affiliation{%
    \institution{State Key Laboratory of Blockchain and Security, Zhejiang University}
    \institution{Hangzhou High-Tech Zone (Binjiang) Institute of Blockchain and Data Security}
  \city{Hangzhou}
  \country{China}
  \postcode{310027}
}

\author{Yunpeng Qing}
\author{Yihe Zhou}
\author{Yuwen Wang}
\author{Mingli Song}
\affiliation{%
    \institution{State Key Laboratory of Blockchain and Security, Zhejiang University}
    \institution{Hangzhou High-Tech Zone (Binjiang) Institute of Blockchain and Data Security}
  \city{Hangzhou}
  \country{China}
}

\renewcommand{\shortauthors}{Feiyang Xu et al.}

\begin{abstract}
{
Active Voltage Control~(AVC) on the Power Distribution Netwo- rks~(PDNs) aims to stabilize the voltage levels to ensure efficient and reliable operation of power systems. With the increasing integration of distributed energy resources, recent efforts have explored employing multi-agent reinforcement learning~(MARL) techniques to realize effective AVC. Existing methods mainly focus on the acquisition of short-term AVC strategies, \textit{i.e.}, only learning AVC within the short-term training trajectories of a singular diurnal cycle. However, due to the dynamic nature of load demands and renewable energy, the operation states of real-world PDNs may exhibit significant distribution shifts across varying timescales~(\textit{e.g.}, daily and seasonal changes). This can render those short-term strategies suboptimal or even obsolete when performing continuous AVC over extended periods. In this paper, we propose a novel temporal prototype-aware learning method, abbreviated as TPA, to learn time-adaptive AVC under short-term training trajectories. At the heart of TPA are two complementary components, namely multi-scale dynamic encoder and temporal prototype-aware policy, that can be readily incorporated into various MARL methods. The former component integrates a stacked transformer network to learn underlying temporal dependencies at different timescales of the PDNs, while the latter implements a learnable prototype matching mechanism to construct a dedicated AVC policy that can dynamically adapt to the evolving operation states. Experimental results on the AVC benchmark with different PDN sizes demonstrate that the proposed TPA surpasses the state-of-the-art counterparts not only in terms of control performance but also by offering model transferability. Our code is available at \url{https://github.com/Canyizl/TPA-for-AVC}.
}

\end{abstract}



\maketitle

\section{Introduction}

Low-carbon energy technologies have made an attractive development path for addressing sustainability concerns in power systems~\cite{first}. Among these technologies, rooftop solar photovoltaics (PVs) in Power Distribution Networks~(PDNs) are particularly prominent, commonly installed atop residential and commercial structures to harness solar energy~\cite{pv}. However, the widespread adoption of PVs has introduced greater variability and intermittency in power generation, thereby exacerbating the voltage fluctuations and posing challenges to the reliability and stability of power systems~\cite{intropp}. Thus, modern PDNs necessitate advanced Active Voltage Control (AVC) schemes that can dynamically adapt to rapid changes in PV generation and load conditions, ensuring voltage profiles within permissible limits using the controllable apparatuses~(\textit{e.g.}, PV inverters and static var compensators)~\cite{apt1,apt2, liu2024MAM}.

To stabilize the voltage levels, the conventional AVC strategy on PDNs involves solving constrained optimization problems. Such optimization methods can be mainly categorized into two classes: Optimal Power Flow~(OPF)-based methods~\cite{opf,opf2} and droop control~(DC)-based ones~\cite{droopc1,droopc2}. OPF-based methods propose to compute an optimal set of operating points for the entire power system that adheres to power balance constraints. In spite of its effectiveness, a key limitation of OPF is its dependency on an exact model of the power system, which can be difficult to obtain. Additionally, the complexity of OPF often results in time-consuming optimization processes, making it challenging to apply in real-time scenarios~\cite{opfweak}.
On the other hand, the DC-based methods are typically based on pre-defined parameters that are manually set. While this method can be implemented more rapidly than OPF, it is generally considered to be sub-optimal due to its reliance on these fixed parameters, which also limits its adaptability across different systems~\cite{droopc}.

To alleviate these issues, deep Multi-Agent Reinforcement Learning~(MARL), a data-driven learning paradigm, has emerged as a promising AVC solution for PDNs due to its rapid response time and effective coordination capabilities. Existing MARL approaches typically begin by partitioning the PDN into multiple regions according to its tree-like structure. This allows individual agents to be assigned the task of AVC within each region~\cite{Two-stage,OldMADDPG1}.
While these region-based approaches have demonstrated effectiveness in smaller-scale PDNs, they may suffer from limitations when applied to larger networks due to the inconsistent regional structures~\cite{MAPDN,TMAAC}.
Thus, the pioneering work of \citet{MAPDN} suggests a shift in focus from regional control to the management of individual distributed energy resources,
which has shown promise in more complex tasks.~Furthermore, to capture the varying significance of nodes within large-scale PDNs, a more recent work~\cite{TMAAC} incorporates the transformer network~\cite{transformer} to enhance the representation ability of~agents.

Despite the promising results achieved, existing works mainly emphasize the development of short-term AVC strategies, \textit{i.e.}, only learning AVC within the short-term training trajectories of a singular diurnal cycle. They often slice the original continuous long-term annual data into short-term daily segments for training. However, PDNs are dynamically evolving systems across varying timescales, exhibiting significant temporal distribution shifts in real-world operation states. 
It can be observed that the temporal distribution shifts are not only influenced by the habits of load demands (substantial disparities between nighttime and daytime) ~\cite{opf}, but also maintain a stable offset with seasonal changes (increased photovoltaic generation in summer compared to winter)~\cite{summer1,summer2}.
Thus, employing short-term strategies directly for continuous AVC over extended durations may lead to unsatisfactory performance.
Even with long-term data for agent training, the model capacity is insufficient to adaptively represent the entire dynamic characteristics of the PDNs. Additionally, longer trajectories necessitate higher computational resources, making it impractical when extending trajectories to hundreds of times.

In this paper, we introduce a novel temporal prototype-aware learning method, termed as TPA, to learn time-adaptive AVC under short-term training trajectories. The proposed TPA comprises two key components, namely multi-scale dynamic encoder and temporal prototype-aware policy, which can be seamlessly integrated with various MARL methods to tackle the challenges arising from the temporal distribution shifts.
Specifically, the multi-scale dynamic encoder employs a stacked transformer network to integrate the minute-level temporal observation with season-level task guidance, which enables the agents to capture temporal dependencies across different timescales within the PDNs.
To derive the final temporal prototype-aware AVC policy, we propose a learnable prototype matching mechanism that realizes time-adaptive adjustments under various operation states.
The prototypes are constrained by several customized losses to serve as representative temporal patterns for diverse climate impacts, thereby offering corresponding decision support.
Our contributions can be summarized as follows:


\begin{itemize}[leftmargin=*]
\item We delve into the challenge of learning the long-term dynamic control policy under short-term training trajectories, a highly practical yet largely overlooked problem in previous data-driven studies of power system control applications.
\item We propose a novel Temporal Prototype-Aware~(TPA) learning method to learn time-adaptive AVC on the PDNs. By integrating the multi-scale dynamic encoder and the temporal prototype-aware policy, TPA enables agents to make decisions that dynamically adapt to the temporal distribution shifts.
\item Extensive experiments on the Bus-141 and Bus-322 PDN benchmarks showcase that the proposed TPA is readily applicable to various MARL methods and yields results superior to the state-of-the-art counterparts in both controllable rate and power generation loss, especially in the long-term operation cycles. Moreover, the additional temporal prototype analysis demonstrates the transferability of TPA across different PDN sizes.
\end{itemize}

\section{Related Works}

\subsection{AVC Methods on PDNs}

Traditional optimization constraint methods for active voltage control~(AVC) can be roughly classified into two categories: 
the optimal power flow~(OPF)-based methods~\cite{opf,opf2} and the droop control~(DC)-based methods~\cite{droopc1,droopc2}. OPF-based methods are regarded as an optimization problem minimizing the total power loss under diverse voltage constraints. \citet{opf2} minimizes the power loss while fulfilling reactive power limits and bus voltage limits to balance power constraints. This method provides the optimal AVC performance if given the exact system model, which is always hard to obtain.
DC-based methods establish a piece-wise linear relationship between PV generation and voltage. This relationship can be utilized to regulate the local bus without acquiring total voltage divisions. DC-based methods~\cite{droopc,droopc1} rely on their manually-designed parameters of the linear relationship, which also limits its adaptability across different systems.
As the PDN scale increases, it is difficult for both of these two methods to react to the rapid environmental changes when facing the instability of new energy.

Recently, the data-driven deep MARL technique~\citep{liu2023CIA,liu2024OPT} has drawn broad attention in addressing AVC task~\citep{OldMADDPG1,OldTD3,MAPDN} due to the fast response and effective interaction ability.
Existing MARL methods for AVC can be broadly classified into two categories: region-based methods~\cite{OldMADDPG1,Two-stage,OldTD3} and PV-based methods~\cite{MAPDN,TMAAC}.
The region-based methods partition the PDN into multiple regions according to its tree-like structure and treat each region as an individual agent to interact with environments. 
Then common MARL algorithms can be integrated to solve the AVC tasks.
\citet{OldMADDPG1} applies MADDPG~\cite{MADDPG} on each region agent to control the reactive power of PV inverters. 
To regular regional reactive power, \citet{Two-stage} combines the OPF-based method to \textrm{MADDPG} improving regional control accuracy.
\citet{OldTD3} utilizes spectral clustering algorithms to formulate the control relationship between regions agent and then applies \textrm{MATD3}~\cite{MATD3} on them, which provides a significant reduction in communication requirements.
While these region-based approaches have demonstrated great effectiveness in smaller-scale PDNs, they suffer from the inconsistent regional number of PVs and inconsistent regional topology when applied to larger networks~\cite{MAPDN}.
Thus PV-based methods have emerged to avoid inconsistent regional structures by modeling each PV inverter as an agent.
\citet{MAPDN} enable PV agents within the same region to share their regional observation, which has demonstrated promise in large-scale benchmarks.
Recently, \citet{TMAAC} introduce the transformer~\cite{transformer} to capture varying significances of every PV in large-scale PDNs, enabling effective representations for the characteristics of the power network.

However, all of these methods focus on the short-term strategy while ignoring the dynamicity of PDNs under varying timescales.
This oversight results in the final suboptimal strategies over extended periods. 
Instead, our TPA enables the agents to be actively aware of the temporal dependencies under different timescales.

\subsection{Prototype Learning}

The prototype learning technique~\cite{proto,proto2,hu2024improving} enables the deep model to generate output based on a small number of human-understandable instances. 
These instances, referred to as prototypes, represent prominent features that capture essential characteristics of different classes or groups. They make significant influences on the final decision while enhancing the representation capability and explainability of deep models~\cite{protgnn, prototowards}.
Prototype learning has been widely explored across diverse fields.
In the image recognition task, \citet{proto} dissect the image by identifying prototypical parts and integrating evidence from these prototypes to generate the final classification, while \citet{proto2} shared the global attribute prototypes among different classes to better extend intermediate features to unseen images.
Meanwhile in the text recognition field, \citet{textpro1} propose prototype trajectories under RNN backbones to emulate human text analysis, identifying the most similar prototype for each sentence.
For further intuitive and fine-grained interpretation, \citet{textpro2} selects sub-sequences of different concepts as prototype parts, which are further compared with the text input to understand model decisions.
As for the reinforcement learning field, recent works introduce visual observation prototype~\cite{protox,protoRL,dreamerpro} to further extract effective representations of observations.
\citet{protoRL} pre-train prototypes using observation data from multiple tasks to extract the summarization of exploratory experiences. The prototypes then further formulate an intrinsic reward to incentivize the exploration of uncharted regions within the state space.

To augment the interpretability of RL decision-making processes in visual games, \citet{protox} employs contrastive learning to generate prototypes identified as key scenarios within the task.  The state inputs are compared with these prototypes to compute a similarity measure for explanation and determine the final action output.
\citet{dreamerpro} learn visual prototypes from the observations, which significantly discards the task-irrelevant visual distractions for decision-making.
Unlike previous methods in RL that passively partition the state space and extract prototypes, our work actively considers temporal priors that potentially influence the strategy, maintaining global temporal prototypes served as representative temporal patterns reflecting various climate impacts.

\section{PRELIMINARY}

\subsection{AVC Problem on PDNs}

In this paper, the power distribution networks (PDNs) installed with roof-top photovoltaics~(PVs) are divided into multiple regions based on its realistic tree topology $\mathbf{G} = \left( \mathcal{V},\mathcal{E} \right) $, where $\mathcal{V} = \left\{ 0,1,...,N_v \right\}$ and $\mathcal{E} = \left\{ 0,1,...,N_e \right\}$ indicate the set of buses~(nodes) and the set of branches~(edges), respectively~\cite{tree}. For each bus $i \in \mathcal{V}$, $p_i + j q_i$ represents the complex power injection, where $p_i$ is active power and $q_{i}$ is reactive power. There are complex and non-linear relationships that satisfy the power system dynamic rules~\cite{MAPDN}.

\begin{figure}[!t]
    \centering                    
    \includegraphics[width=0.7\columnwidth]{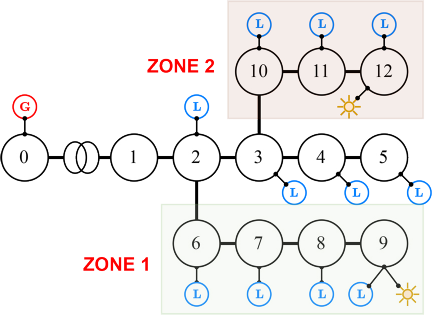}  
    \caption{An example of the PDN. Each bus is indexed by a circle with a number. ``$\mathbf{G}$" denotes the external generator. ``$\mathbf{L}$" denotes loads. ``sun" denotes the location of a PV installed. We control the voltages on bus 2--12. Bus 0--1 represents the main system with the constant voltage outside the PDN.}
    \label{fig:pp}
\end{figure}

As shown in Figure.~\ref{fig:pp}, different control regions are divided according to their shortest path from the terminal bus to the main branch~\cite{MAPDN}. Each node carries its own load, and some nodes also carry PV. In our problem, each PV is considered as an agent and controls the voltage around a stationary value denoted as $v_{ref}$. For safe operation of the distribution network, 5\% voltage fluctuation is usually under consideration. Specifically, let the standard value $v_{ref} = 1.00$ per unit~($p.u.$). Voltage amplitude of each bus needs to satisfy $0.95\,p.u.\leq v_{ref}\leq 1.05\,p.u.,\,\forall{i} \in \mathcal{V} \textbackslash \left\{ 0 \right\}$.
Some extreme situations may lead to exceeding safety thresholds.
For example, the end-user voltage could be smaller than $0.95\,p.u.$ when the load is heavy during the nighttime, whereas the solar energy during summer may cause the voltage amplitude to exceed $1.05\,p.u.$.

\subsection{Dec-POMDP of AVC}
\label{sec:dec}

\begin{figure*}[!t]
    \centering                            
    \includegraphics[width=0.9\textwidth]{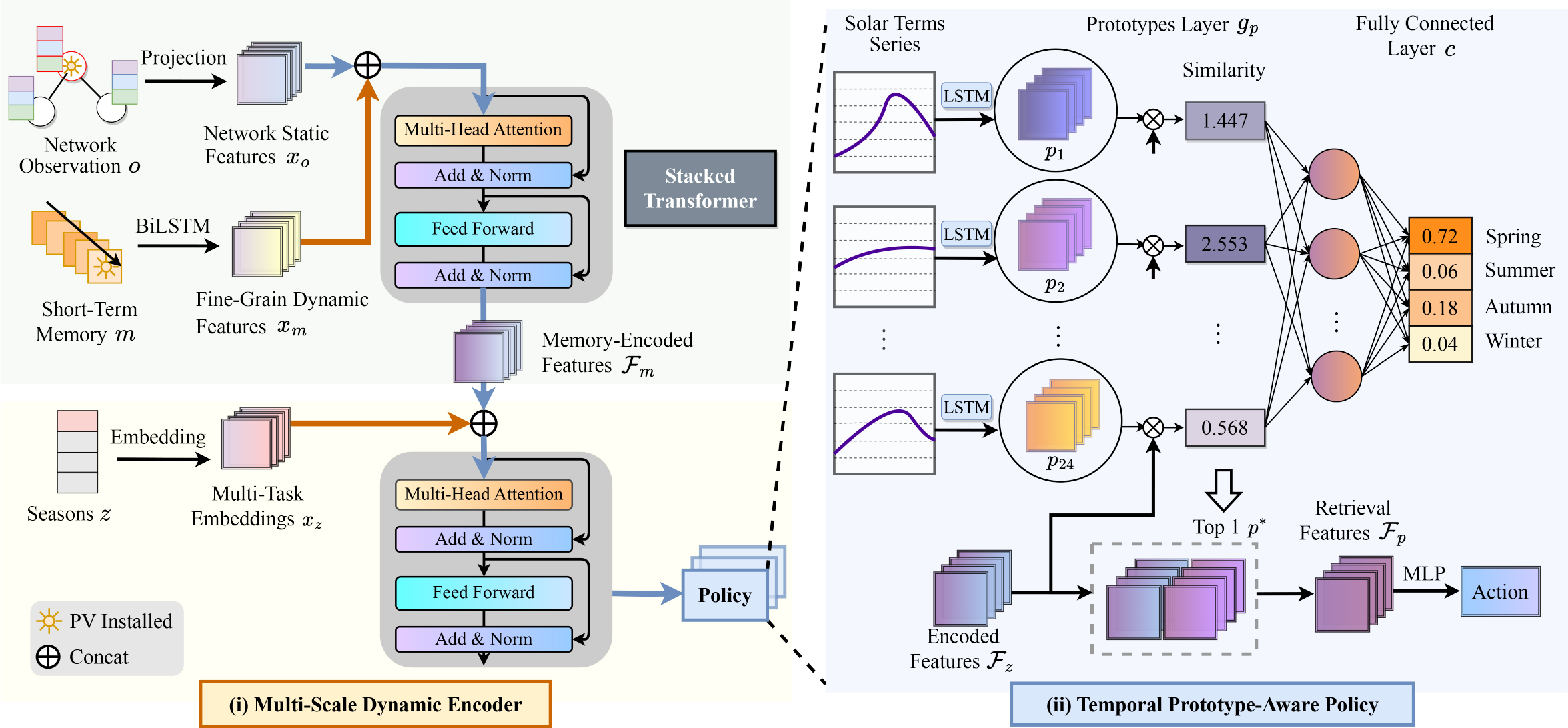} 
    \caption{ Illustration of the proposed method. The module~(i) is employed for extracting multi-scale temporal dependencies while module~(ii) constructs a prototype matching mechanism to enable agents to dynamically adjust their strategies.
} 
    \label{fig:model} 
\end{figure*}

We define the cooperation among PV agents as MARL tasks, further
formulated as a Dec-POMDP~\cite{Dec}.
A formal Dec-POMDP is formulated as a tuple such that $ \left( \mathcal{I}, \mathcal{S}, \mathcal{A}, \mathcal{O}, \mathcal{T}, r, \Omega, \mathcal{\gamma} \right) $. The components of the problem definition are described in detail as follows:

\textbf{Agents:} $\mathcal{I} = \left\{ 1,...,n \right\}$ denotes the set of $n$ agents. In the AVC, we treat every PV invert as an agent.

\textbf{State and Observation:} $\mathcal{S}$ is the state set and $\mathcal{O}$ is the observation set, where the observation $o_i$ of agent $i$ is drawn from the observation function $\Omega(s,i)$. In this problem, all agents~(\textit{i.e.}, PVs) can only observe partial information from their own region instead of the global state of the power system. Because the regional PV ownership may be independent and separated~\cite{seppv}. The observation set is defined as $\mathcal{O} = \mathcal{D} \times \mathcal{P} \times \mathcal{R} \times \mathcal{M}$, where $\mathcal{D} = \left\{ (\boldsymbol{p}^{\ell},\boldsymbol{q}^{\ell}) | {\ell \in \mathcal{V}_{\ell}} \right\}$ is a set of active and reactive powers of loads. $\mathcal{V}_{\ell}$ is the set of all load nodes. $\mathcal{P} = \left\{ (\boldsymbol{p}^{pv},\boldsymbol{q}^{pv}) | {pv \in \mathcal{V}_{pv}}  \right\}$ is a set of current active powers and the preceding reactive powers. $\mathcal{V}_{pv}$ is the set of all PV nodes. $\mathcal{R} = \left\{ (v_{\ell},\omega_{\ell}) | {\ell \in \mathcal{V} {\ell}} \right\}$ is a set of voltage wherein $v_{\ell}$ is a vector of voltage magnitudes and $\omega_{\ell}$ is a vector of voltage phases measured in radius. $\mathcal{M}$ is the predefined zone index for each bus.

\textbf{Action:} Each agent $i \in \mathcal{I}$ is equipped with a continuous action set $\mathcal{A}$, which represents the ratio of reactive power it generates. The transition probability $\mathcal{T} : \mathcal{S} \times \mathcal{A}^n \times \mathcal{S} \rightarrow [0,1]$ denotes any state $s \in \mathcal{S}$ to any state $s' \in \mathcal{S}$ after taking action $\mathbf{a} \in \mathcal{A}^n$. 

\textbf{Reward:} The reward function is defined as follows:
\begin{equation}
\label{eq:rf}
    r = - \frac{1}{|\mathbf{V}|}\sum_{i \in \mathbf{V}}\ell_v(v_i) \, - \, \alpha \cdot \ell_q(\boldsymbol{q}^{pv}),
\end{equation}
where $\ell_v(\cdot)$ is a voltage barrier function~(detailed in Appendix~\ref{sec:apx_da}). $\ell_q(\boldsymbol{q}^{pv}) = \frac{1}{|\mathcal{N}|}||\boldsymbol{q}^{{pv}}||_1$ is the reactive power generation loss. $\alpha$ is a hyper-parameter to balance the $\ell_v(\cdot)$ and $\ell_q(\boldsymbol{q}^{pv})$ on networks of various scales. The $\ell_v(\cdot)$ loss measures the voltage deviations whether in the safe range.
$\ell_q(\boldsymbol{q}^{pv})$ controls the reactive power generation as little as possible to reduce the power waste. We denote the return as $\sum_{k=0}^{\infty} \gamma^{k}r_{k+1}$, where $\mathcal{\gamma} \in [0,1)$ is the discount factor.

\section{Method}
In this section, we further detail the proposed  Temporal
Prototype-Aware learning~(TPA) method.
As shown in Figure~\ref{fig:model}, TPA consists of two core modules, namely multi-scale dynamic encoder and temporal prototype-aware policy.
In the former module, we introduce a stacked transformer network to encode input features with short-term memory and seasonal labels, capturing the temporal dependencies. 
In the latter module, the encoded features are employed to derive the prototype learning with temporal patterns of season climates and offer tailored decision support. 

\subsection{Multi-scale Dynamic Encoder}
Due to the dynamic nature of load demands and renewable energy, the operation states of real-world PDNs may exhibit significant distribution shifts across varying timescales~(\textit{e.g.}, daily and seasonal changes).
The goal of the multi-scale dynamic encoder is to take network states as input and enhance them with both minute-level temporal observation and season-level task guidance, training the stacked transformer network that accomplishes extracting multi-scale underlying temporal dependencies for each PV agent. 

The observation of agent $i$ at time step $t$ is denoted as $o_i^t \in \mathcal{O}$, which contains features of buses in the same region. For simplicity, we omit the index $i$ of the agent in the subsequent section. We extract network representations from $o^t$ via a projection layer $g_o$.  The obtained network static feature $x_o = g_o(o^t) \in \mathbb{R}^{r \times h}$ represents the regional network features at $t$ step, 
where $r$ is the number of buses within the same region
and $h$ is the latent dimension. To introduce temporal dependencies to static features, we collect the period of $K$ steps as a short-term memory $m^t = [q^{pv}_{t - K}, ..., q^{pv}_{t - 1} ]$, where $q^{pv}_{t-k}$ denotes the previous reactive power $q^{pv}$ of the agent at $t-k$ step. The short-term memory $m^t$ intuitively reflects the current trend at a minute level, so we use a bidirectional long short-term memory model $g_m$ to extract the fine-grain dynamic features $x_m=g_m(m) \in \mathbb{R}^{r \times h}$.

To capture temporal dependencies from static and dynamic inputs, we use the stacked transformer network~\cite{transformer} to obtain encoded features layer by layer. As depicted in the left column of Figure~\ref{fig:model}, we use transformer $f(\cdot;\psi): \mathbb{R}^{r \times 2h} \rightarrow \mathbb{R}^{2h}$ parameterized by $\psi$ to encode essential static and dynamic features with dimension $2h$ following the general attention mechanism as:
\begin{equation}
\begin{split}
    Q^{(\ell)},K^{(\ell)},V^{(\ell)} &= \left[ W_Q^{(\ell-1)}, W_K^{(\ell-1)}, W_V^{(\ell-1)} \right] \cdot E^{(\ell-1)},  \\
    \bar{Y}^{(\ell)} &= \textrm{softmax}\left( \frac{Q^{l} {K^{(\ell)}}^{\top}}{\sqrt{d_k}} \cdot V^{(\ell)} \right), \\
    E^{(\ell)} &= \textrm{LayerNorm}\left( E^{(\ell-1)} + W_{Y}·\bar{Y}^{(\ell)} \right),
\end{split}
\label{eq:tslayer}
\end{equation}
where $W_Q, W_K, W_V$ and $W_Y$ represent learnable parameters. $d_k$ is a scaling factor used to adjust the scale of attention scores. 
$E^{\ell} \in \mathbb{R}^{2h}$ are the encoded features with dimension $2h$ computed after $\ell$ steps of the transformer network. The input features $E^{0}$ is initialized using the pair of latent inputs $[x_i,x_j]$, where $x_i$ denotes the static-based inputs and $x_j$ denotes the dynamic inputs. After running $L$ iterations of Eq.~(\ref{eq:tslayer}), the stacked transformer network can generate the final encoded features as $\mathcal{F} = E^{(L)} \in \mathbb{R}^{2h}$. Here we adopt two transformer branches to capture two temporal dependencies.


In the fine-grain branch, we concatenate inputs $[x_o, x_{m}]$ and feed to the stacked transformer network, focusing on capturing latent regional features enhanced with fine-grain temporal dependencies. The transformer generates memory-encoded features as:
\begin{equation}
\mathcal{F}_m = f([x_{o}, x_{m}];\psi_m) \in \mathbb{R}^{2h}.
\end{equation}
After the fine-grain branch, we get the one-hot season labels $z$ at $t$ step, which is clearly shared by all agents. We use an embedding layer $g_z$ to obtain multi-task embeddings $x_z=g_z(z) \in \mathbb{R}^{r \times h}$ to guide the large variation of temporal dependencies during different seasons under every region. To concatenate $\mathcal{F}_m$ and $\boldsymbol{I}_s$, we implement another linear layer $g_h$ to reduce dimensions of $\hat{\mathcal{F}}_m = g_h(\mathcal{F}_m): \mathbb{R}^{2h} \rightarrow \mathbb{R}^{h}$. Then we feed the season-level inputs $[\hat{\mathcal{F}}_m,x_z]$ to the course-gain branch of the stacked transformer network, further introducing broad temporal dependencies on the foundation of memory-enhanced network features. The transformer generates final encoded features as:
\begin{equation}
\mathcal{F}_z = f([\hat{\mathcal{F}}_m,x_z];\psi_z) \in \mathbb{R}^{2h}.
\end{equation}
Here we employ the transformer to learn network representations from static and dynamic input pairs. The multi-scale dynamic encoder can capture the multi-scale temporal dependencies from the raw observations and temporal features. The encoded features $\mathcal{F}_{z}$ are further employed by the policy module.



\subsection{Temporal Prototype-Aware Policy}

Existing methods suffer from that are easily suboptimal or even obsolete when performing continuous AVC over extended periods. The policy should dynamically adapt to the evolving operation states. At the heart of the temporal prototype-aware policy is global temporal patterns of season climates, which enables the agent to explicitly perceive season climates.

Drawing inspiration from effective load forecasting through 24 solar terms~\cite{solarterms}, we partitioned each season into 6 adaptable prototypes, resulting in 24 temporal prototypes over a year.
We feed the whole day data~(one prototype corresponds to one day) into LSTM to initialize temporal prototypes $\mathcal{P} = \left\{ p_1,p_2,...,p_{24} \mid p_i \in \mathbb{R}^{h} \right\}$. To minimize additional data as much as possible, we select three major power states $x_{p} \in \mathbb{R} ^{d \times 3}$~(where $d$ denotes the real-time control period number on the whole day): the active and reactive powers of loads, and the active powers of PVs to initialize the temporal prototypes. 
We match the highest similarity prototypes $p^*$ to make the policy be aware of the current season climate, following the calculation of similarity:
\begin{equation}
\label{eq:sim}
   \textrm{sim}(p_i, \mathcal{F}_z) = \log \left( \frac{||p_i - \mathcal{F}_z||^2_2 + 1} {||p_i - \mathcal{F}_z||^2_2 + \epsilon}\right), 
\end{equation}
where $\epsilon$ is set to a small value preventing division by zero. The linear layer $g_c$ with parameters $\phi$ is employed to dynamically adapt encoded features to evolving climate features, extracting the retrieval features $\mathcal{F}_p = g_c\left( 
\mathcal{F}_z,p^*;\phi \right) \in \mathbb{R}^{h}$. We use multilayer perceptrons~(MLPs) as the action prediction network $g_a$ to predict the logits of action in continuous control $a = g_a(\mathcal{F}_p) $, which is the action in MARL interacting with environments.

The objective function of prototype learning $\mathcal{L}_{pl}$ is defined as:
\begin{equation}
    \mathcal{L}_{pl} = \mathcal{L}_{ce}(c\, \circ   \, g_p \circ  \mathcal{F}_z, z) + \lambda_{1} \mathcal{L}_{clst} + \lambda_{2} \mathcal{L}_{sep} + \lambda_{3} \mathcal{L}_{div},
\label{eq:pl}
\end{equation}
where $c$ is a fully connected layer that predicts season probability and $g_p$ is the prototype layer.
$\mathcal{L}_{ce}$ represents the cross-entropy loss for season classification and $z_s$ is season labels. $\lambda_{1},\lambda_{2},\lambda_{3}$ are hyper-parameters controlling the weights of losses. 

\begin{figure*}[!t]
  \centering
    \begin{subfigure}{1.0\textwidth}
    \centering
    \includegraphics[width=1.0\textwidth]{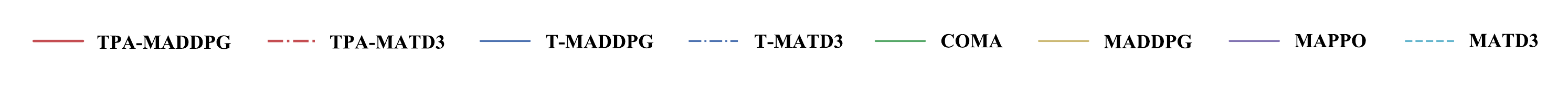}
  \end{subfigure}

    \begin{subfigure}{0.16\textwidth}
    \centering
    \includegraphics[width=\textwidth]{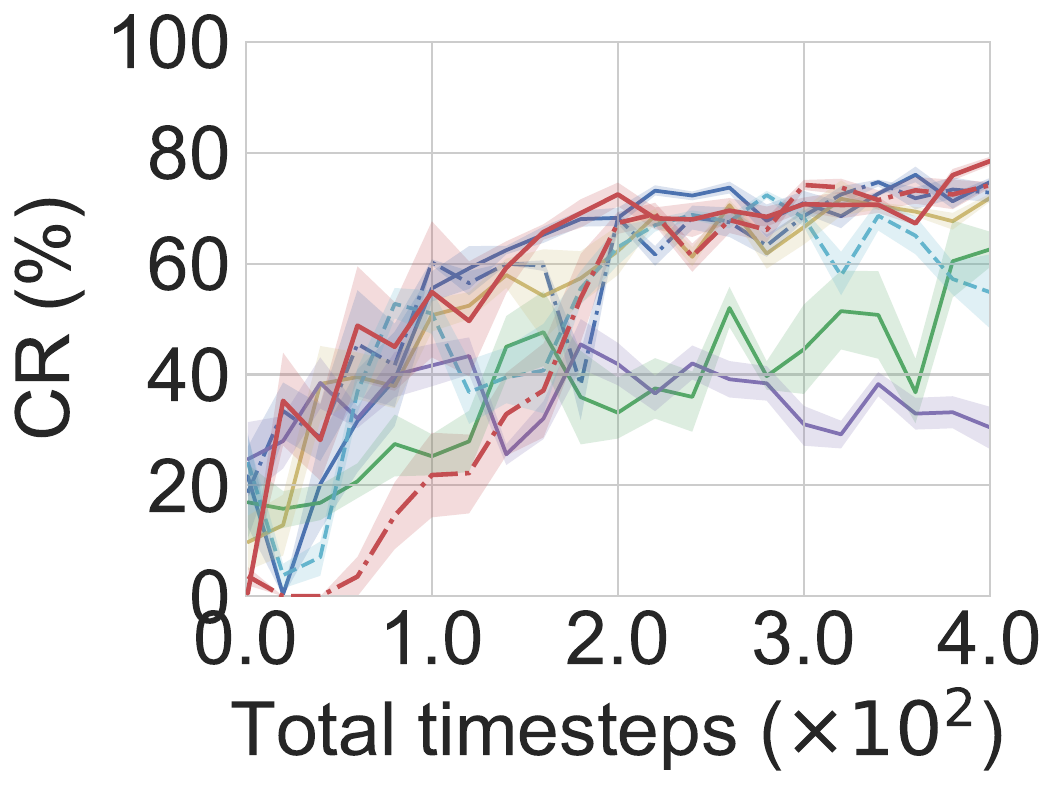}
    \caption{CR-L1-322}
  \end{subfigure}
  \begin{subfigure}{0.16\textwidth}
    \centering
    \includegraphics[width=\textwidth]{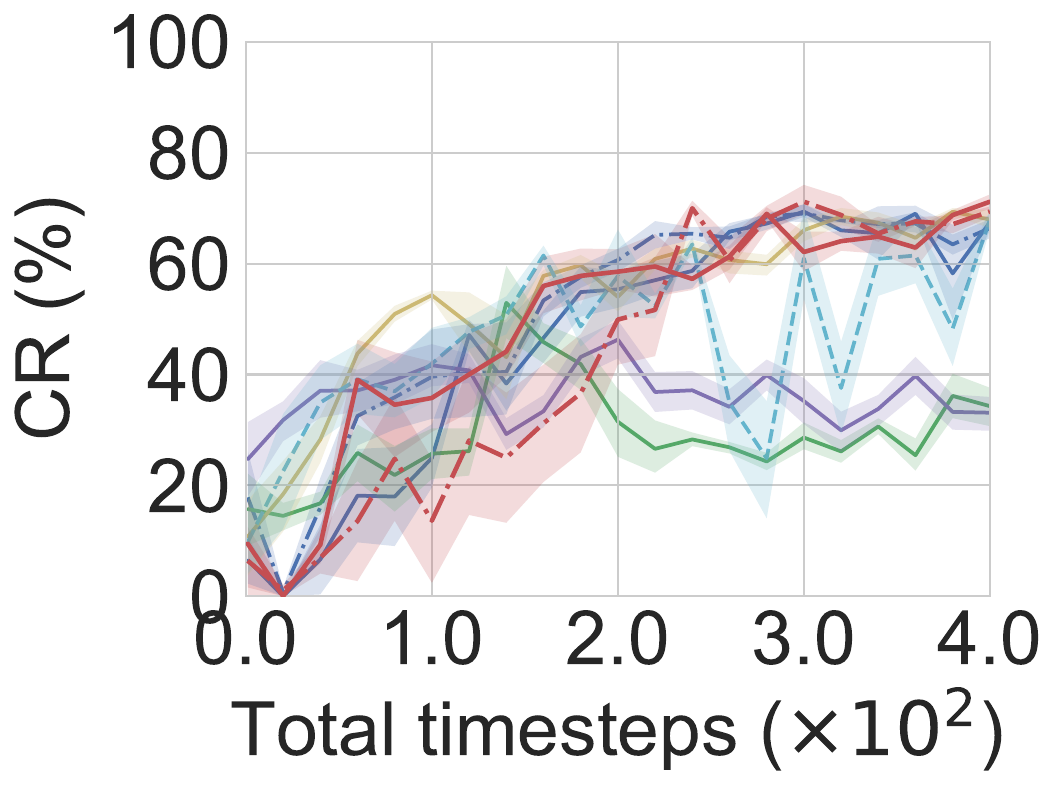} 
    \caption{CR-BOWL-322}
  \end{subfigure}
      \begin{subfigure}{0.16\textwidth}
    \centering
    \includegraphics[width=\textwidth]{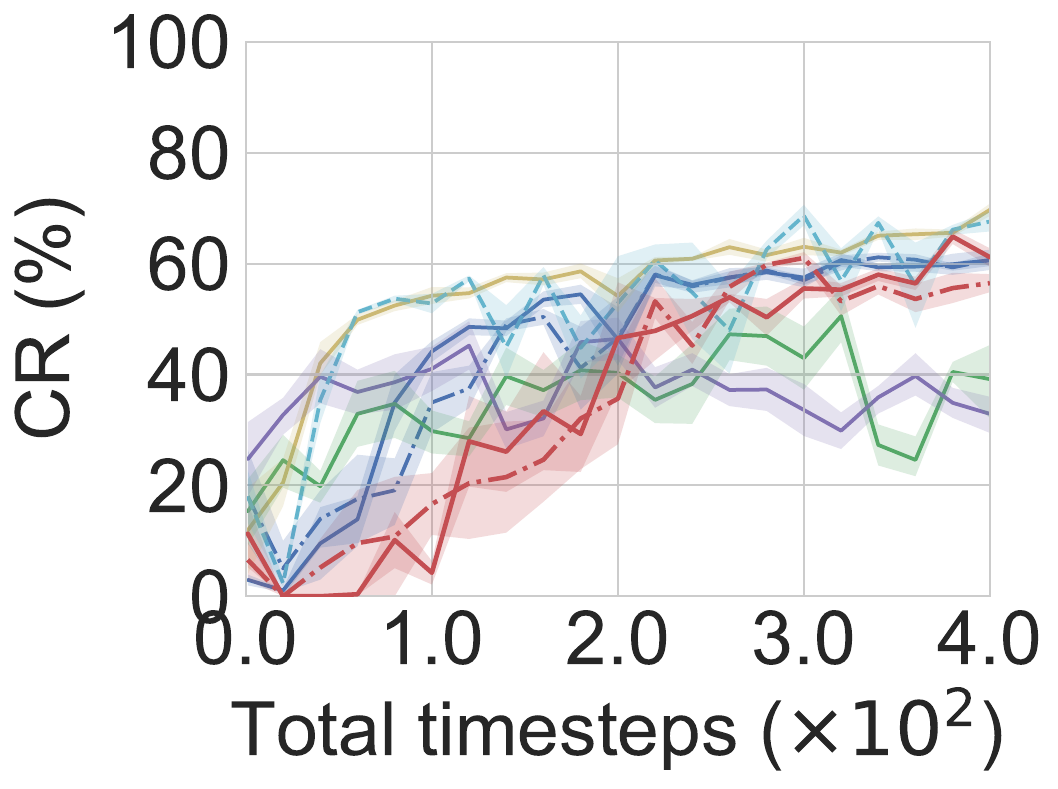}
    \caption{CR-L2-322}
  \end{subfigure}
      \begin{subfigure}{0.16\textwidth}
    \centering
    \includegraphics[width=\textwidth]{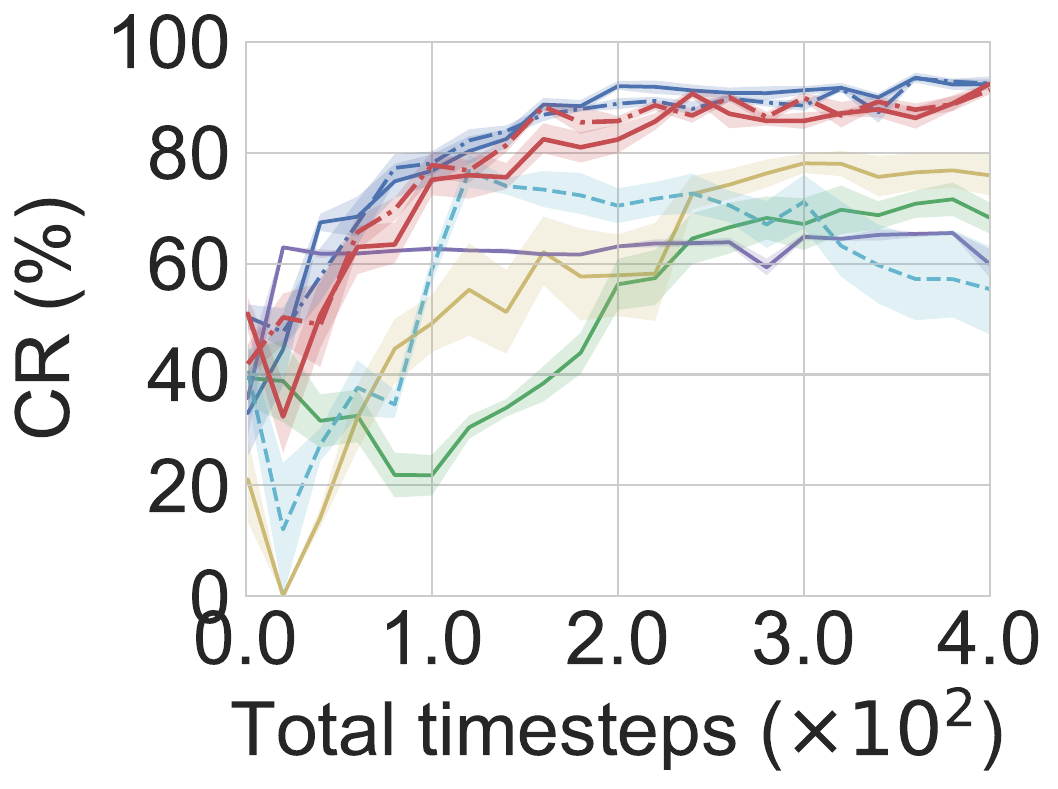}
    \caption{CR-L1-141}
  \end{subfigure}
  \begin{subfigure}{0.16\textwidth}
    \centering
    \includegraphics[width=\textwidth]{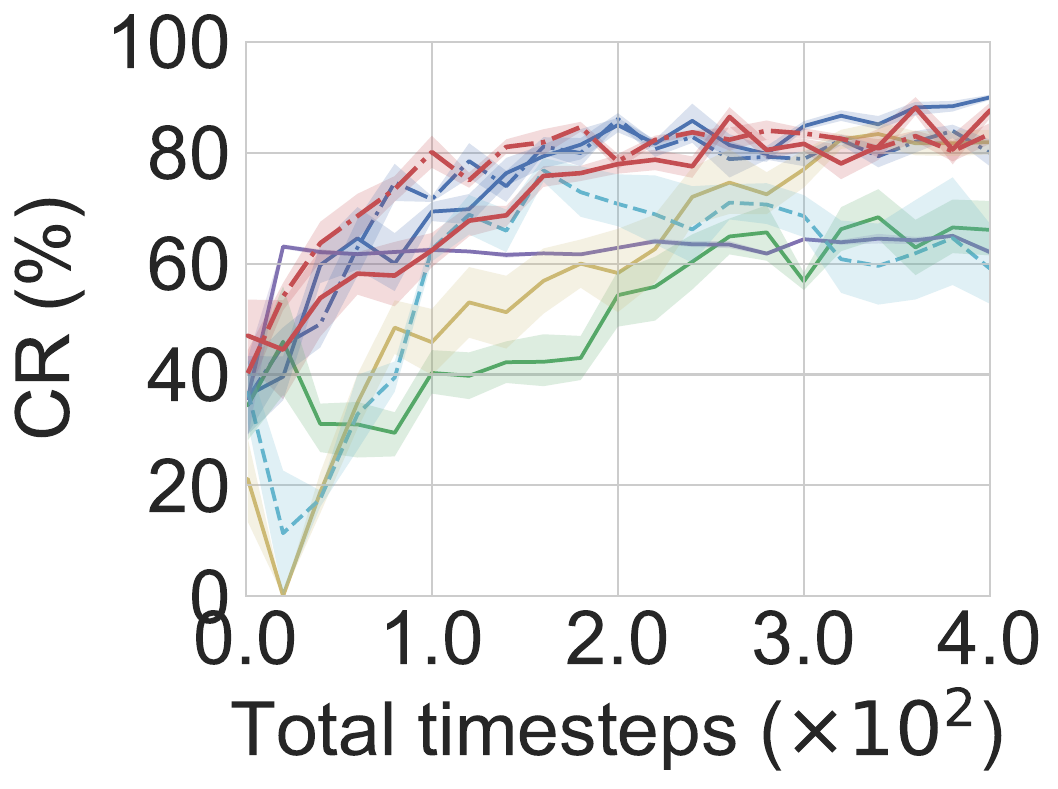} 
    \caption{CR-BOWL-141}
  \end{subfigure}
      \begin{subfigure}{0.16\textwidth}
    \centering
    \includegraphics[width=\textwidth]{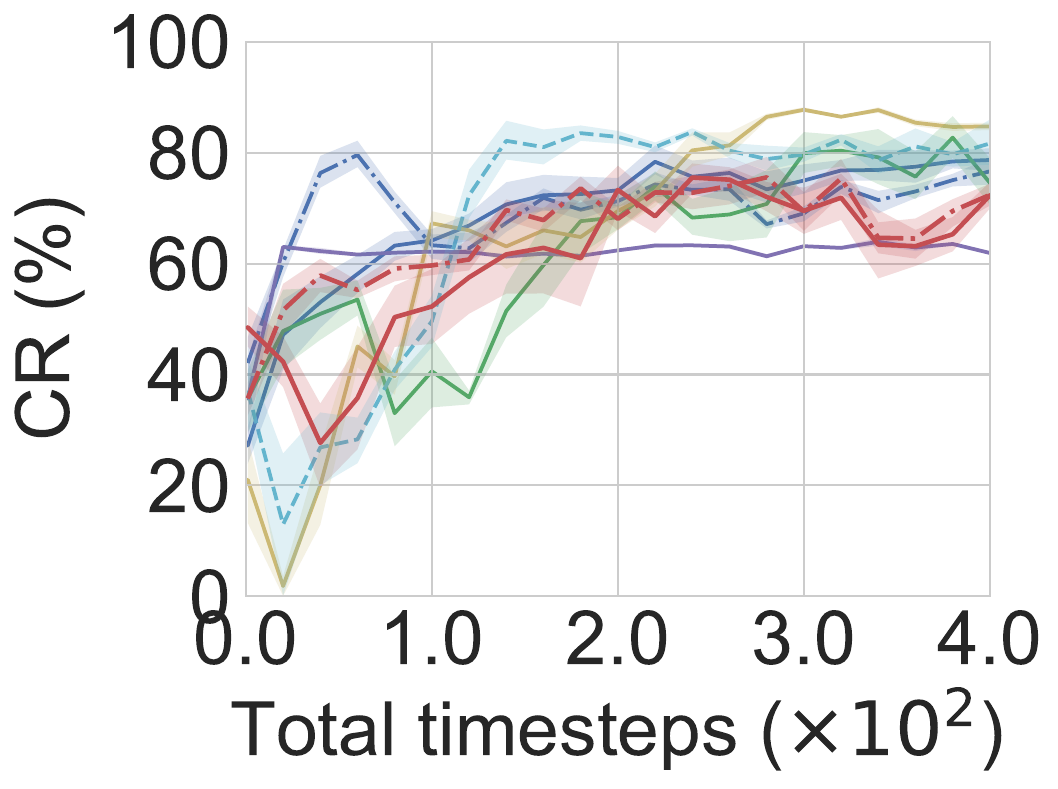}
    \caption{CR-L2-141}
  \end{subfigure}
  
  \begin{subfigure}{0.16\textwidth}
    \centering
    \includegraphics[width=\textwidth]{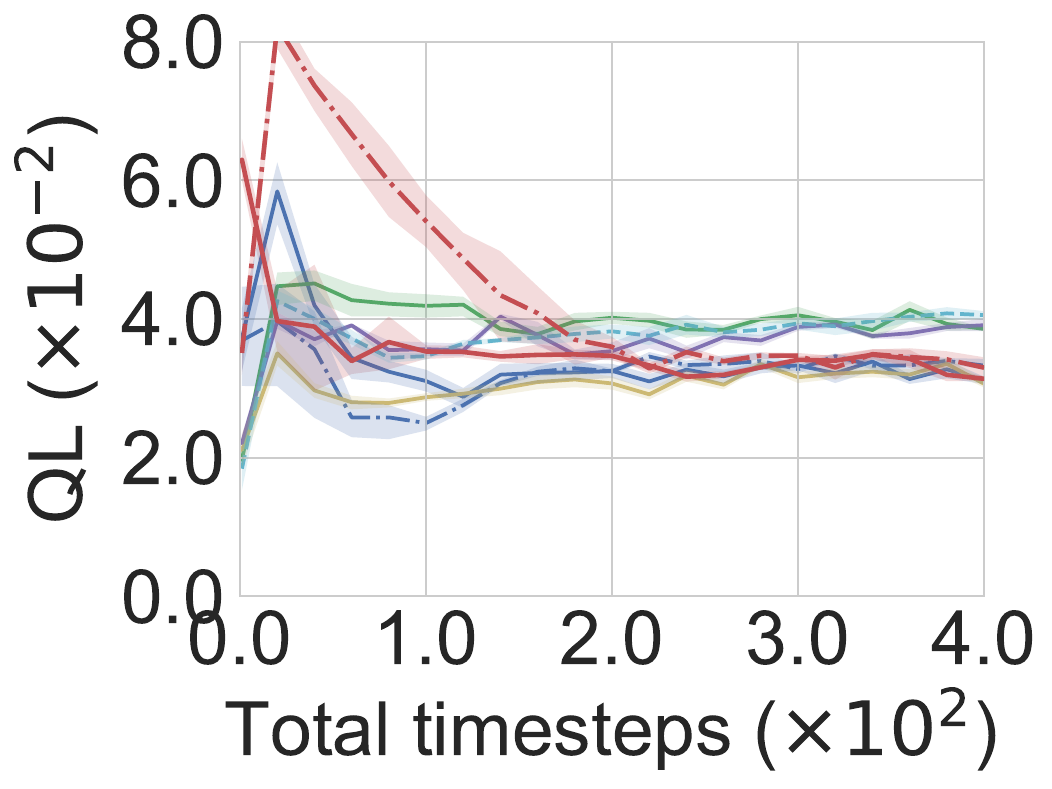} 
    \caption{QL-L1-322}
  \end{subfigure}
      \begin{subfigure}{0.16\textwidth}
    \centering
    \includegraphics[width=\textwidth]{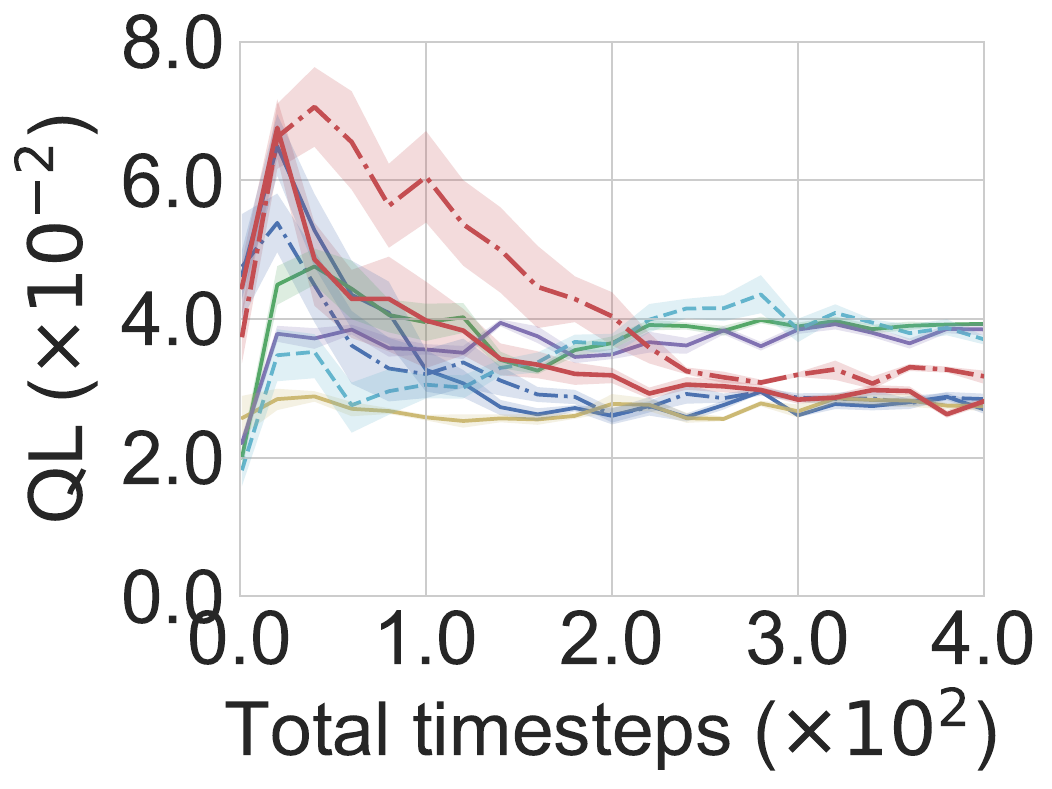}
    \caption{QL-BOWL-322}
  \end{subfigure}
  \begin{subfigure}{0.16\textwidth}
    \centering
    \includegraphics[width=\textwidth]{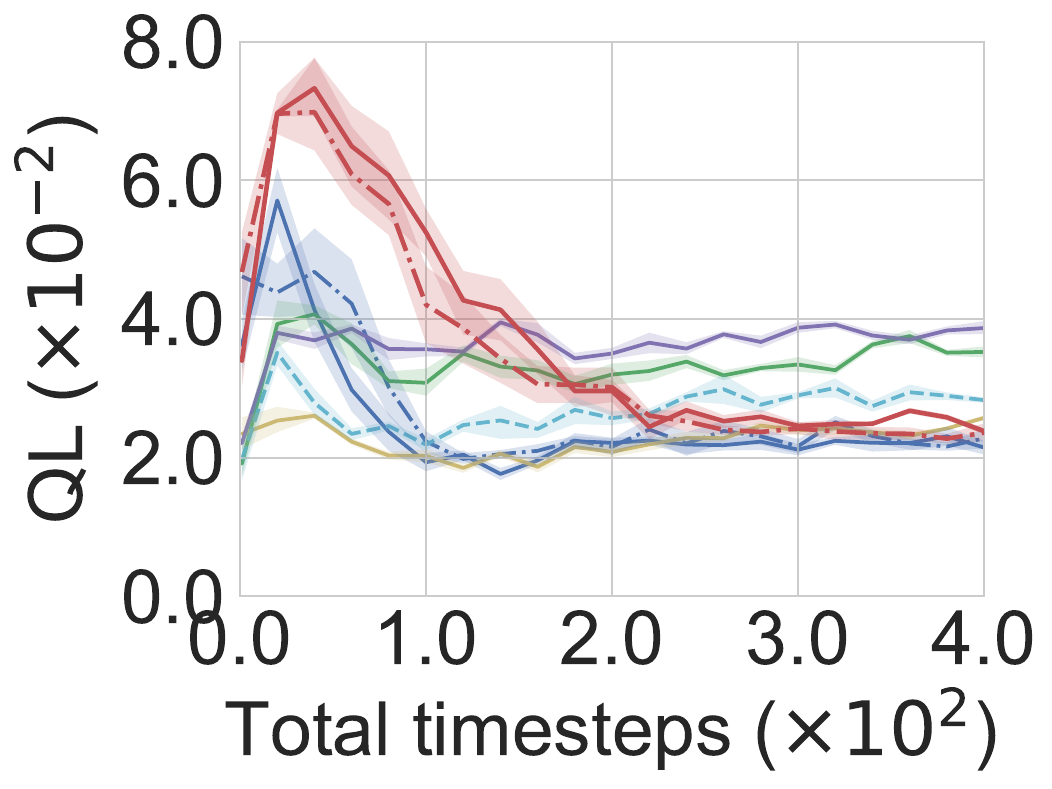} 
    \caption{QL-L2-322}
  \end{subfigure}
  \begin{subfigure}{0.16\textwidth}
    \centering
    \includegraphics[width=\textwidth]{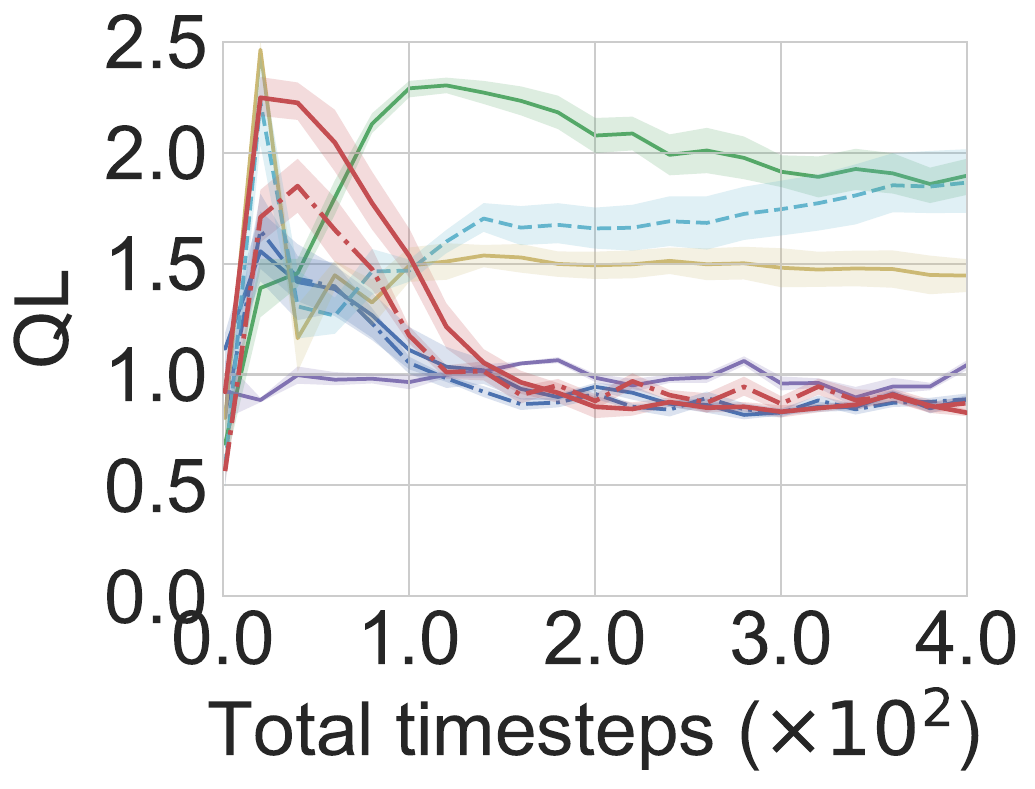} 
    \caption{QL-L1-141}
  \end{subfigure}
      \begin{subfigure}{0.16\textwidth}
    \centering
    \includegraphics[width=\textwidth]{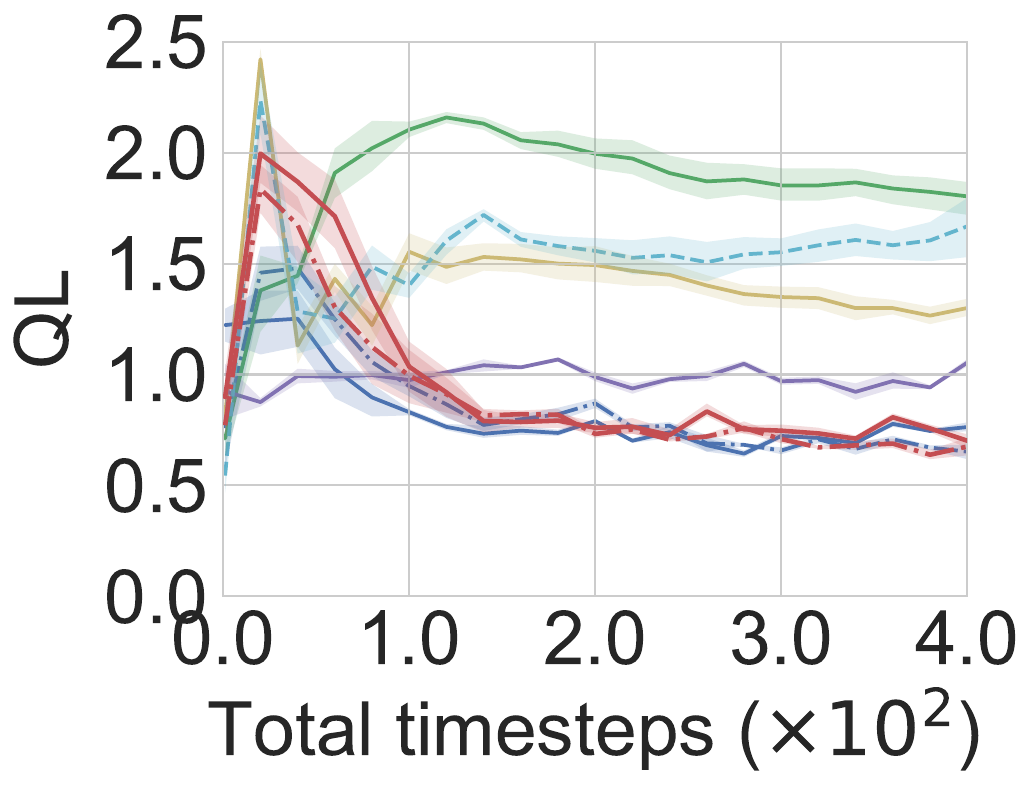}
    \caption{QL-BOWL-141}
  \end{subfigure}
  \begin{subfigure}{0.16\textwidth}
    \centering
    \includegraphics[width=\textwidth]{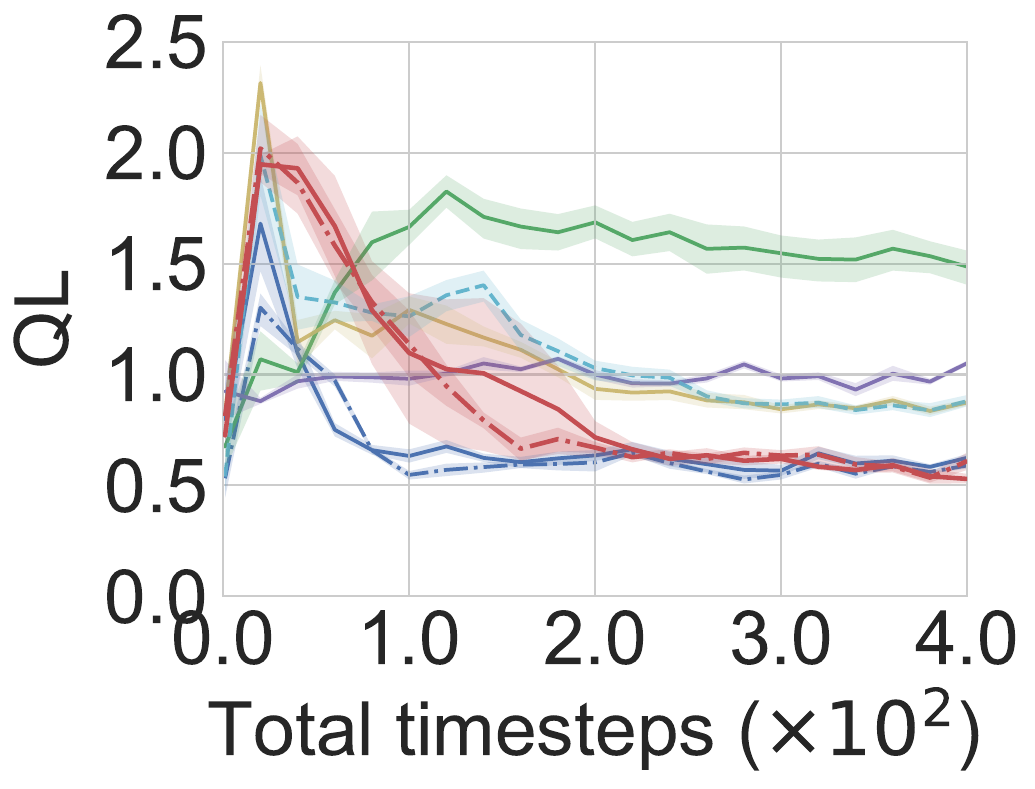} 
    \caption{QL-L2-141}
  \end{subfigure}
  \caption{Median CR and QL of algorithms with different voltage barrier functions. The sub-caption indicates metric-Barrier-scenario.  ``TPA-'' refers to the combination of our framework with other methods, while ``T-'' represents the incorporation of the previous TMAAC with other methods. All experimental results are illustrated with the mean and the standard deviation of the metrics over 5 random seeds for a fair comparison. To make the results clearer for readers, we adopt a 50\% confidence interval to plot the error region.}
  \label{fig:sota}
\end{figure*}

\begin{table*}[!t]
\centering
\caption{Test results of algorithms in 322-bus and 141-bus system. 
\textbf{Bold} denotes the best results. 
$\pm$ corresponds to the standard deviation of testing episodes.
}
\label{tab:sota}
\resizebox{0.98\textwidth}{!}{
\begin{tabular}{@{}ll|cccccccccc@{}}
\toprule
\multicolumn{2}{c|}{\multirow{2}{*}{\textbf{Methods}}} & \multicolumn{2}{c}{\textbf{Spring}}                                 & \multicolumn{2}{c}{\textbf{Summer}}                                 & \multicolumn{2}{c}{\textbf{Fall}}                                   & \multicolumn{2}{c}{\textbf{Winter}}                                & \multicolumn{2}{c}{\textbf{Average}}        \\ \cmidrule(l){3-12} 
\multicolumn{2}{c|}{}                         & \textbf{CR(\%)}               & \multicolumn{1}{c|}{\textbf{QL($\frac{\textrm{MW}}{\textrm{MVAR}}$)}}           & \textbf{CR(\%)}               & \multicolumn{1}{c|}{\textbf{QL($\frac{\textrm{MW}}{\textrm{MVAR}}$)}}           & \textbf{CR(\%)}               & \multicolumn{1}{c|}{\textbf{QL($\frac{\textrm{MW}}{\textrm{MVAR}}$)}}           & \textbf{CR(\%)}              & \multicolumn{1}{c|}{\textbf{QL($\frac{\textrm{MW}}{\textrm{MVAR}}$)}}           & \textbf{CR(\%)}        & \textbf{QL($\frac{\textrm{MW}}{\textrm{MVAR}}$)}           \\ \midrule
\multicolumn{2}{l|}{322-MAPPO}                & 57.5{\scalebox{0.8}{$\pm$9.4}}           & \multicolumn{1}{c|}{0.035}          & 41.2{\scalebox{0.8}{$\pm$5.8}}           & \multicolumn{1}{c|}{0.035}          & 56.1{\scalebox{0.8}{$\pm$11.2}}          & \multicolumn{1}{c|}{0.035}          & 78.7{\scalebox{0.8}{$\pm$17.8}}         & \multicolumn{1}{c|}{0.036}          & 58.4          & 0.035          \\
\multicolumn{2}{l|}{322-COMA}                 & 82.4{\scalebox{0.8}{$\pm$18.3}}          & \multicolumn{1}{c|}{0.038}          & 70.3{\scalebox{0.8}{$\pm$17.2}}          & \multicolumn{1}{c|}{0.038}          & 83.4{\scalebox{0.8}{$\pm$18.1}}          & \multicolumn{1}{c|}{0.038}          & 96.9{\scalebox{0.8}{$\pm$5.3}}          & \multicolumn{1}{c|}{0.038}          & 83.3          & 0.038          \\
\multicolumn{2}{l|}{322-MATD3}                & 87.0{\scalebox{0.8}{$\pm$17.4}}          & \multicolumn{1}{c|}{0.038}          & 74.3{\scalebox{0.8}{$\pm$17.3}}          & \multicolumn{1}{c|}{0.039}          & 89.2{\scalebox{0.8}{$\pm$17.9}}          & \multicolumn{1}{c|}{0.038}          & 99.0{\scalebox{0.8}{$\pm$4.5}}          & \multicolumn{1}{c|}{0.038}          & 87.4          & 0.038          \\
\multicolumn{2}{l|}{322-MADDPG}               & 86.2{\scalebox{0.8}{$\pm$18.1}}          & \multicolumn{1}{c|}{0.032}          & 72.0{\scalebox{0.8}{$\pm$18.6}}          & \multicolumn{1}{c|}{0.033}          & 87.5{\scalebox{0.8}{$\pm$18.4}}          & \multicolumn{1}{c|}{0.032}          & 98.3{\scalebox{0.8}{$\pm$5.7}}          & \multicolumn{1}{c|}{0.031}          & 86.0          & 0.032          \\
\multicolumn{2}{l|}{322-T-MADDPG}             & 91.2{\scalebox{0.8}{$\pm$14.8}}          & \multicolumn{1}{c|}{0.031}          & 82.5{\scalebox{0.8}{$\pm$21.9}}          & \multicolumn{1}{c|}{0.033}          & 92.9{\scalebox{0.8}{$\pm$13.1}}          & \multicolumn{1}{c|}{0.031}          & 97.8{\scalebox{0.8}{$\pm$7.8}}          & \multicolumn{1}{c|}{\textbf{0.029}} & 91.1          & 0.031          \\
\multicolumn{2}{l|}{322-T-MATD3}              & 93.5{\scalebox{0.8}{$\pm$13.0}}          & \multicolumn{1}{c|}{0.030}          & 87.5{\scalebox{0.8}{$\pm$21.3}}          & \multicolumn{1}{c|}{\textbf{0.029}} & 95.4{\scalebox{0.8}{$\pm$12.2}}          & \multicolumn{1}{c|}{\textbf{0.027}} & 97.7{\scalebox{0.8}{$\pm$7.8} }         & \multicolumn{1}{c|}{0.030}          & 93.5          & \textbf{0.029} \\ \midrule
\multicolumn{2}{l|}{\cellcolor{gray!13}322-TPA-MATD3}            & 93.2{\scalebox{0.8}{$\pm$12.8} }         & \multicolumn{1}{c|}{0.032}          & 86.3{\scalebox{0.8}{$\pm$21.8}  }        & \multicolumn{1}{c|}{0.031}          & 93.3{\scalebox{0.8}{$\pm$13.4}   }       & \multicolumn{1}{c|}{0.030}          & 97.2{\scalebox{0.8}{$\pm$6.6}     }     & \multicolumn{1}{c|}{0.030}          & 92.5          & 0.030          \\
\multicolumn{2}{l|}{\cellcolor{gray!13}322-TPA-MADDPG}           & \textbf{95.1}{\scalebox{0.8}{$\pm$11.5}} & \multicolumn{1}{c|}{\textbf{0.029}} & \textbf{89.9}{\scalebox{0.8}{$\pm$21.0} } & \multicolumn{1}{c|}{\textbf{0.029}} & \textbf{97.1}{\scalebox{0.8}{$\pm$10.4} } & \multicolumn{1}{c|}{0.029}          & \textbf{99.1}{\scalebox{0.8}{$\pm$3.1} } & \multicolumn{1}{c|}{\textbf{0.029}} & \textbf{95.3 } & \textbf{0.029} \\ \midrule
\multicolumn{2}{l|}{141-MAPPO}                & 76.3{\scalebox{0.8}{$\pm$20.2}}          & \multicolumn{1}{c|}{1.025}          & 61.0{\scalebox{0.8}{$\pm$13.9}}          & \multicolumn{1}{c|}{0.946}          & 78.4{\scalebox{0.8}{$\pm$23.0}}          & \multicolumn{1}{c|}{1.027}          & 97.2{\scalebox{0.8}{$\pm$6.2}}          & \multicolumn{1}{c|}{1.148}          & 78.2          & 1.037          \\
\multicolumn{2}{l|}{141-COMA}                 & 87.1{\scalebox{0.8}{$\pm$4.8}}           & \multicolumn{1}{c|}{1.631}          & 86.3{\scalebox{0.8}{$\pm$9.9}}           & \multicolumn{1}{c|}{1.707}          & 88.3{\scalebox{0.8}{$\pm$4.1}}           & \multicolumn{1}{c|}{1.631}          & 87.2{\scalebox{0.8}{$\pm$3.5}}          & \multicolumn{1}{c|}{1.554}          & 87.2          & 1.631          \\
\multicolumn{2}{l|}{141-MADDPG}               & 97.1{\scalebox{0.8}{$\pm$6.9}}           & \multicolumn{1}{c|}{1.140}          & 96.6{\scalebox{0.8}{$\pm$11.3}}          & \multicolumn{1}{c|}{1.105}          & 99.4{\scalebox{0.8}{$\pm$3.5}}           & \multicolumn{1}{c|}{1.148}          & 99.8{\scalebox{0.8}{$\pm$0.7}}          & \multicolumn{1}{c|}{1.232}          & 98.2          & 1.156          \\
\multicolumn{2}{l|}{141-MATD3}                & 97.5{\scalebox{0.8}{$\pm$5.7}}           & \multicolumn{1}{c|}{1.134}          & 97.4{\scalebox{0.8}{$\pm$9.3}}           & \multicolumn{1}{c|}{1.151}          & 99.9{\scalebox{0.8}{$\pm$0.7}}           & \multicolumn{1}{c|}{1.140}          & \textbf{100}{\scalebox{0.8}{$\pm$0.0}}  & \multicolumn{1}{c|}{1.144}          & \textbf{98.7} & 1.142          \\
\multicolumn{2}{l|}{141-T-MATD3}              & 97.0{\scalebox{0.8}{$\pm$4.3}}           & \multicolumn{1}{c|}{0.816}          & 97.4{\scalebox{0.8}{$\pm$7.9}}           & \multicolumn{1}{c|}{0.836}          & 99.9{\scalebox{0.8}{$\pm$0.8}}           & \multicolumn{1}{c|}{0.810}          & \textbf{100}{\scalebox{0.8}{$\pm$0.0}}  & \multicolumn{1}{c|}{0.836}          & 98.5          & 0.825          \\
\multicolumn{2}{l|}{141-T-MADDPG}             & \textbf{97.4} {\scalebox{0.8}{$\pm$5.1}}  & \multicolumn{1}{c|}{0.877}          & 97.8{\scalebox{0.8}{$\pm$8.8}}           & \multicolumn{1}{c|}{0.871}          & 99.8{\scalebox{0.8}{$\pm$0.6} }          & \multicolumn{1}{c|}{0.877}          & \textbf{100}{\scalebox{0.8}{$\pm$0.0}}  & \multicolumn{1}{c|}{0.938}          & \textbf{98.7} & 0.891          \\ \midrule
\multicolumn{2}{l|}{\cellcolor{gray!13}141-TPA-MATD3}            & 97.1{\scalebox{0.8}{$\pm$4.1}}           & \multicolumn{1}{c|}{0.889}          & 96.8{\scalebox{0.8}{$\pm$7.6}}           & \multicolumn{1}{c|}{0.889}          & 98.5{\scalebox{0.8}{$\pm$0.7}}           & \multicolumn{1}{c|}{0.868}          & \textbf{100}{\scalebox{0.8}{$\pm$0.0}}  & \multicolumn{1}{c|}{0.909}          & 98.1          & 0.888          \\
\multicolumn{2}{l|}{\cellcolor{gray!13}141-TPA-MADDPG}           & 97.1{\scalebox{0.8}{ $\pm$4.2}}           & \multicolumn{1}{c|}{\textbf{0.734} } & \textbf{97.8}{\scalebox{0.8}{$\pm$7.4}}  & \multicolumn{1}{c|}{\textbf{0.793} } & \textbf{99.9}{\scalebox{0.8}{$\pm$0.5}}  & \multicolumn{1}{c|}{\textbf{0.789}} & \textbf{100}{\scalebox{0.8}{$\pm$0.0}}  & \multicolumn{1}{c|}{\textbf{0.772} } & \textbf{98.7} & \textbf{0.772 } \\ \bottomrule
\end{tabular}
}
\end{table*}
As shown in Eq.~(\ref{eq:pl}), several constraints are proposed to construct the final prototypes~\cite{protgnn}.
Firstly, the cluster cost $\mathcal{L}_{clst}$ in Eq.~(\ref{eq:clst}) motivates encoded features to exhibit proximity to one prototype corresponding to their season.
\begin{equation}
\mathcal{L}_{clst} =  \frac{1}{n}\sum_{i=1}^{\mathcal{N}} \min_{j:p_j\in \mathcal{P}_{y_i}} ||\mathcal{F}_z - p_j||^{2}_2,
\label{eq:clst}
\end{equation}
where $\mathcal{P}_{y_i}$ is the set of prototypes under $y_i$ class.
Secondly, the separation cost $\mathcal{L}_{sep}$ in Eq.~(\ref{eq:sep}) promotes the distancing of encoded features from prototypes that do not belong to their seasons.
\begin{equation}
\mathcal{L}_{sep} = - \frac{1}{n}\sum_{i=1}^{\mathcal{N}} \min_{j:p_j \notin \mathcal{P}_{y_i}} ||\mathcal{F}_z - p_j||^{2}_2.\\
\label{eq:sep}
\end{equation}
Finally, the diversity loss $\mathcal{L}_{div}$ in Eq.~(\ref{eq:div}) encourages the diversity of the learned prototypes via penalizing prototypes too close.
\begin{equation}
\mathcal{L}_{div} = \sum_{k=1}^{4} \sum_{\substack{i \neq j \\ p_i, p_j \in \mathcal{P}_k}} \max \left(0, \cos\left(p_i,p_j\right) - \xi \right),
\label{eq:div}
\end{equation}
where $k = \left\{1,2,3,4 \right\}$ denotes 4 seasons. $\xi$ is the threshold of the cosine similarity in the diversity loss.


For MARL process, there are $n$ agents with policies $\boldsymbol{\mu} = \{{\mu}_{\theta_1},...,{\mu}_{\theta_n}\}$, where each policy is parameterized by $\theta_i$ all proceed through the subsequent steps. At each control step $t$, the agent $i$ with a given observation $o_i$ under state $s$ will select the action $a_i$ with exploration noises and storage to the experience replay buffer $\mathcal{B}$. Then the agent will receive shared rewards $r$ according to the reward function in Eq.~(\ref{eq:rf}) and the next observation $s'$. The global Q-function $Q_{global}^{\boldsymbol{\mu}}$ embedded the value of actions. The joint agents with policies can be optimized by our total objective function $\mathcal{L}_{pl} + \mathcal{L}_{ac}$. The objective function of prototype learning $\mathcal{L}_{pl}$ is formulated in Eq.~(\ref{eq:pl}). We perform deterministic policy gradient~\cite{MADDPG} over  $\mathcal{L}_{ac}$ as:
\begin{equation}
\label{eq:policy}
\nabla_{\theta_i}\mathcal{L}_{ac} = \vmathbb{E}_{s,a \sim \mathcal{B}} \left[\nabla_{\theta_i} {\mu_i}(a_i|o_i) \nabla_{a_i} Q_{global}^{\boldsymbol{\mu}}(s,a_1,...,a_n) \mid_{ a_i= {\mu}_i(o_i)}  \right].
\end{equation}
Then the critic network estimates the long-term impact of actions compared with real rewards, which are optimized by
\begin{equation}
\begin{split}
\mathcal{L}_{critic} & = \vmathbb{E}_{s,a,r,s'} \left[ \left(Q_{global}^{\boldsymbol{\mu}}(s,a_1,...,a_n) - y \right), ^ 2 \right],\\
y & = r + \gamma Q_{global}^{\boldsymbol{\mu}'}(s',a_1',...,a_n')\mid_{ a_i'= \boldsymbol{\mu}'(o_i)}.
\end{split}
\end{equation}
where $\boldsymbol{\mu}' = \{ {\mu}_{\theta_1'},...,{\mu}_{\theta_n'} \}$ is the set of target policies with delayed parameters $\theta_i'$.
In addition, the proposed TPA serves as a plug-and-play module readily applicable to various MARL algorithms including MADDPG~\cite{MADDPG} and MATD3~\cite{MATD3}.

\section{Experiments}
To illustrate the effectiveness of the proposed TPA method, we conduct experiments on the MAPDN benchmark~(\citet{MAPDN}). We aim to answer the following questions: 
(1) Can TPA outperform the state-of-the-art MARL methods on both the singular diurnal cycle and longer cycles?~(Section~\ref{sec:sdt} and Section~\ref{sec:lct}) 
(2) How do different components of TPA contribute to the overall performance?~(Section~\ref{sec:abl}) 
(3) Can the TPA offer the prototypes transferability?~(Section~\ref{sec:trans})
Besides, the visualization analysis is given in Section~\ref{sec:va}.

\subsection{Experimental Settings}
\label{sec:expset}

\textbf{Tasks and Datasets.} 
We consider the active voltage control task on 141-bus and 322-bus power distribution networks with large scale in MAPDN benchmark~\citep{MAPDN}.
The 141-bus network is divided into 9 zones and contains 84 loads and 22 PVs, while the 322-bus network is divided into 22 zones and contains 337 loads and 38 PVs. 
The main objective of the 141 network setting is reducing power waste losses, while the 322 network setting aims to promote control rates.
The reward function is defined by Eq.~(\ref{eq:rf}), where the weight $\alpha$ assigned to $\ell_q(\boldsymbol{q}^{\textrm{pv}})$ in the reward function balances the control stability and power generation wastes on networks of various scales. 
The load and PV data, extracted from three years of real-world data, are interpolated at a 3-minute resolution to match the real-time control period and further randomly initialize the network scenarios. 
During training, we randomly sample the initial state for an episode and each episode lasts for 240 time steps (i.e. a half day)~\cite{MAPDN}.
More details of datasets and implementation are provided in Appendix~\ref{sec:apx_da}. 

\textbf{Evaluation Metrics.} 
We utilize the Controllable Rate~(CR) and Q Loss~(QL)  to qualitatively evaluate the performance following~\citet{MAPDN}.
CR quantifies the effectiveness by measuring the ratio of time steps during which all bus voltages are under control. QL indicates the mean reactive power generations by agents per time step to assess the power waste.

\textbf{Comparison methods.} 
We compare the proposed TPA to several state-of-the-art MARL methods in continuous action space, including MAPPO~\cite{MAPPO}, COMA~\cite{COMA}, MADDPG~\cite{MADDPG}, MATD3~\cite{MATD3}. For a fair comparison, we follow MAPDN~\cite{MAPDN} to strengthen the basic MARL methods by PV-based modeling. Both T-MADDPG and T-MATD3 are proposed in the TMAAC~\cite{TMAAC}, which incorporates the transformer to capture regional dependencies in PDNs.

\subsection{Results on Short-term Operation Cycles}
\label{sec:sdt}

We first compare the performances of our TPA and other baselines under the singular diurnal cycle in active voltage control tasks. Figure~\ref{fig:sota} presents the learning curves of comparison methods on different power networks, while the final best performance is shown in Table~\ref{tab:sota}. The average metrics are used for learning curves.

The results show that TPA can effectively address the AVC problem in the singular diurnal cycle, while also reducing power waste.
Specifically, from the perspective of effective voltage control, the TPA-based architecture significantly outperforms the common one. For example on the 322-bus system, all TPA-based methods achieve over 90\% in the average CR metric, while the common state-of-art method MATD3 only achieves 87.4\%. 
Meanwhile, the quantitative results showcase that the best model of TPA leads the TMAAC ones by 1.8\% in the average CR metric.
The comparison highlights that agents can be effectively enhanced with multi-scale temporal dependencies. 
From the perspective of power loss, TPA achieves comparable and superior CR while exhibiting a low QL value.
For example on the 141-bus system, other methods like MADDPG and MADTD3 exhibit larger QL when achieving the relatively high CR metric in all test cycles.
In contrast, our approach not only attains state-of-the-art on the CR metric but also reaches the minimum on the QL metric.
The superiority of our TPA method can be attributed to the temporal prototypes that effectively utilize seasonal climates and dynamically adapt to evolving operational states.
This approach has resulted in a significant improvement in voltage quality compared to alternative methods.

Meanwhile, TPA exhibits high robustness in extreme situations.  As shown in Table~\ref{tab:sota}, summer is the most challenging season of all year. The CR metric of all methods in summer shows the lowest CR and higher standard deviation in all seasons, due to its intense photovoltaic fluctuations. However, our TPA makes a more significant improvement in the safety of the strategy in summer to achieve high CR and low QL. Meanwhile, the standard deviation of our TPA in other seasons also achieves the lowest value compared to other methods. A detailed visualization analysis of the strategy in four seasons is presented in Appendix~\ref{sec:va}.

\subsection{Results on Long-term Operation Cycles}
\label{sec:lct}

\begin{table}[!t]
\centering
\caption{Test results of our TPA and other baselines in 322-bus and 141-bus systems under longer testing cycles. The average metrics are used for all cycles.}
\resizebox{1.0\columnwidth}{!}{
\begin{tabular}{@{}ll|cccccc@{}}
\toprule
\multicolumn{2}{c|}{\multirow{2}{*}{\textbf{Method}}} & \multicolumn{2}{c}{\textbf{Day}}                             & \multicolumn{2}{c}{\textbf{Month}}                           & \multicolumn{2}{c}{\textbf{Year}}       \\ \cmidrule(l){3-8} 
\multicolumn{2}{c|}{}                        & \textbf{CR(\%)}        & \multicolumn{1}{c|}{\textbf{QL($\frac{\textrm{MW}}{\textrm{MVAR}}$)}}             & \textbf{CR(\%)}        & \multicolumn{1}{c|}{\textbf{QL($\frac{\textrm{MW}}{\textrm{MVAR}}$)}}                                  & \textbf{CR(\%)}        & \textbf{QL($\frac{\textrm{MW}}{\textrm{MVAR}}$)}             \\ \midrule
\multicolumn{2}{l|}{322-MAPPO}               & 58.4          & \multicolumn{1}{c|}{0.035}          & 58.6          & \multicolumn{1}{c|}{0.035}          & 58.6          & 0.035          \\
\multicolumn{2}{l|}{322-COMA}                & 83.3          & \multicolumn{1}{c|}{0.038}          & 81.6          & \multicolumn{1}{c|}{0.038}          & 81.6          & 0.038          \\
\multicolumn{2}{l|}{322-MATD3}               & 87.4          & \multicolumn{1}{c|}{0.038}          & 85.9          & \multicolumn{1}{c|}{0.039}          & 85.7          & 0.039          \\
\multicolumn{2}{l|}{322-MADDPG}              & 86.0          & \multicolumn{1}{c|}{0.032}          & 85.7          & \multicolumn{1}{c|}{0.032}          & 85.5          & 0.032          \\
\multicolumn{2}{l|}{322-T-MADDPG}            & 91.1          & \multicolumn{1}{c|}{0.031}          & 89.3          & \multicolumn{1}{c|}{0.030}          & 88.2          & 0.030          \\
\multicolumn{2}{l|}{322-T-MATD3}             & 93.5          & \multicolumn{1}{c|}{\textbf{0.029}} & 91.5          & \multicolumn{1}{c|}{\textbf{0.028}} & 90.4          & 0.031          \\ \midrule
\multicolumn{2}{l|}{\cellcolor{gray!13}322-TPA-MATD3}           & 92.5          & \multicolumn{1}{c|}{0.030}          & 91.9          & \multicolumn{1}{c|}{0.032}          & \textbf{93.4} & \textbf{0.029} \\
\multicolumn{2}{l|}{\cellcolor{gray!13}322-TPA-MADDPG}          & \textbf{95.3} & \multicolumn{1}{c|}{\textbf{0.029}} & \textbf{93.4} & \multicolumn{1}{c|}{0.030}          & 92.2          & 0.030          \\ \midrule
\multicolumn{2}{l|}{141-MAPPO}               & 78.2          & \multicolumn{1}{c|}{1.037}          & 77.3          & \multicolumn{1}{c|}{1.025}          & 77.0          & 1.025          \\
\multicolumn{2}{l|}{141-COMA}                & 87.2          & \multicolumn{1}{c|}{1.631}          & 85.8          & \multicolumn{1}{c|}{1.642}          & 85.8          & 1.644          \\
\multicolumn{2}{l|}{141-MADDPG}              & 98.2          & \multicolumn{1}{c|}{1.156}          & 96.5          & \multicolumn{1}{c|}{1.161}          & 96.4          & 1.161          \\
\multicolumn{2}{l|}{141-MATD3}               & \textbf{98.7} & \multicolumn{1}{c|}{1.142}          & 97.2          & \multicolumn{1}{c|}{1.154}          & 97.1          & 1.154          \\
\multicolumn{2}{l|}{141-T-MADDPG}            & 98.5          & \multicolumn{1}{c|}{0.825}          & 97.3          & \multicolumn{1}{c|}{0.910}          & 97.2          & 0.910          \\
\multicolumn{2}{l|}{141-T-MATD3}             & \textbf{98.7} & \multicolumn{1}{c|}{0.891}          & 97.7          & \multicolumn{1}{c|}{0.839}          & 97.5          & 0.841          \\ \midrule
\multicolumn{2}{l|}{\cellcolor{gray!13}141-TPA-MATD3}           & 98.1          & \multicolumn{1}{c|}{0.888}          & 97.5          & \multicolumn{1}{c|}{0.904}          & 97.5          & 0.903          \\
\multicolumn{2}{l|}{\cellcolor{gray!13}141-TPA-MADDPG}          & \textbf{98.7} & \multicolumn{1}{c|}{\textbf{0.772}} & \textbf{98.1} & \multicolumn{1}{c|}{\textbf{0.815}} & \textbf{98.0} & \textbf{0.815} \\ \bottomrule
\end{tabular}
}
\label{tab:longer}
\end{table}

To further evaluate our TPA performances under long-term operation cycles, we conduct experiments on the 141-bus and 322-bus networks under 3 different cycles: day, month, and year. Concretely, we still train our TPA with the short-term training trajectories of a half-day, yet assess the performance on longer cycles.
The final performances on average metrics are presented in Table~\ref{tab:longer}.

The results demonstrate that TPA performs with high robustness under longer cycles. 
Specifically, as the test cycle grew longer, the metrics of all other methods showed a decline, which to some extent confirmed our initial idea that longer cycles would bring the problems of multi-scale temporal distribution shifts. 
However, our TPA with robustness on the temporal scale can alleviate this issue and surpass other methods with minimal declines.
When performing continuous AVC over extended periods, the short-term strategies of other methods will deviate from the initial temporal dependencies, gradually moving towards suboptimal solutions. Instead, TPA can dynamically adapt to the evolving operation states and maintain robust control stability even over prolonged periods.

Moreover, the test results under both the singular diurnal cycle and longer cycles show that TPA can effectively serve as a plug-and-play module. 
As shown in Figure~\ref{fig:sota}, two TPA-based MARL algorithms both demonstrate powerful performance.
The TPA-based methods consistently sustain CR surpassing 90\% in longer cycles, outperforming the common MADDPG and MATD3 methods. Especially on the longest year cycle, the TPA-MATD3 method outpaces MATD3 by 7.7\%, and the TPA-MADDPG method surpasses MADDPG by 6.7\%.
The TPA module is readily applicable to various MARL algorithms and enhances their strategies with temporal awareness.

\begin{figure}[!t]
  \centering
    \begin{subfigure}{0.44\textwidth}
    \centering
    \includegraphics[width=0.9\textwidth] {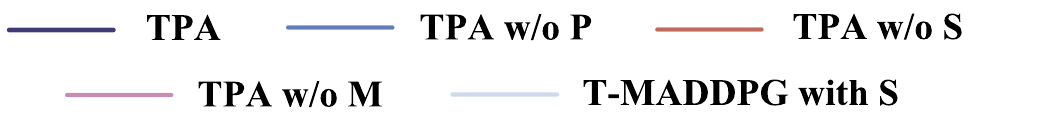}
  \end{subfigure}

  \begin{subfigure}{0.22\textwidth}
    \centering
    \includegraphics[width=\textwidth]{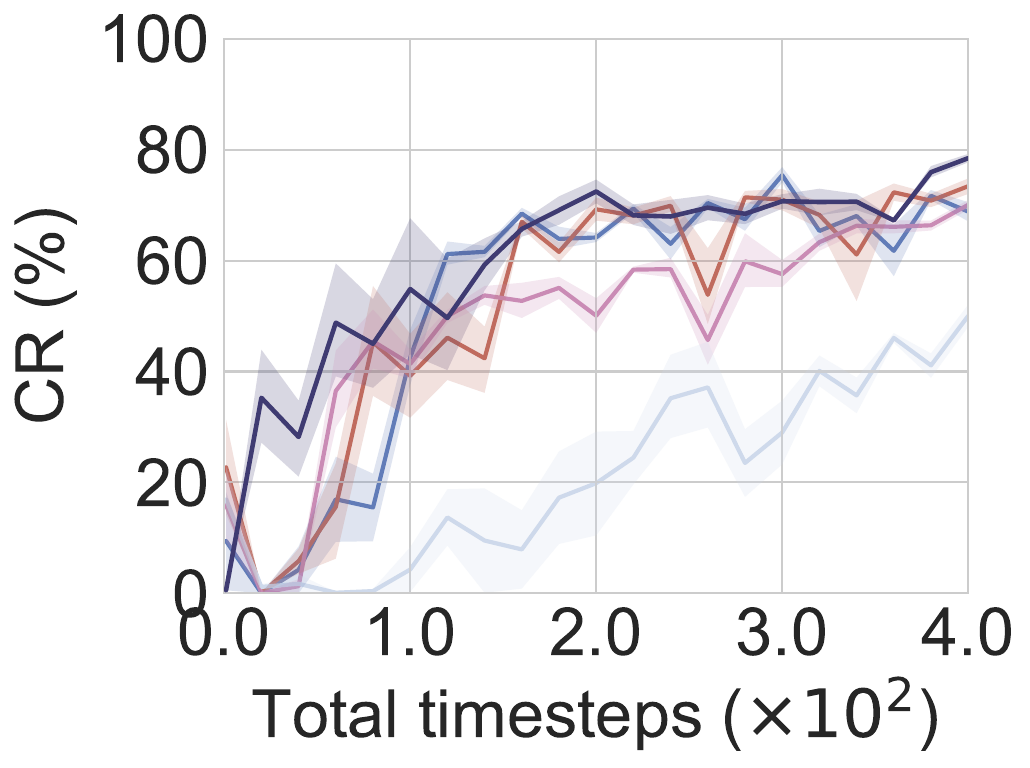}
    \caption{CR-L1-322}
  \end{subfigure}
  \begin{subfigure}{0.22\textwidth}
    \centering
    \includegraphics[width=\textwidth]{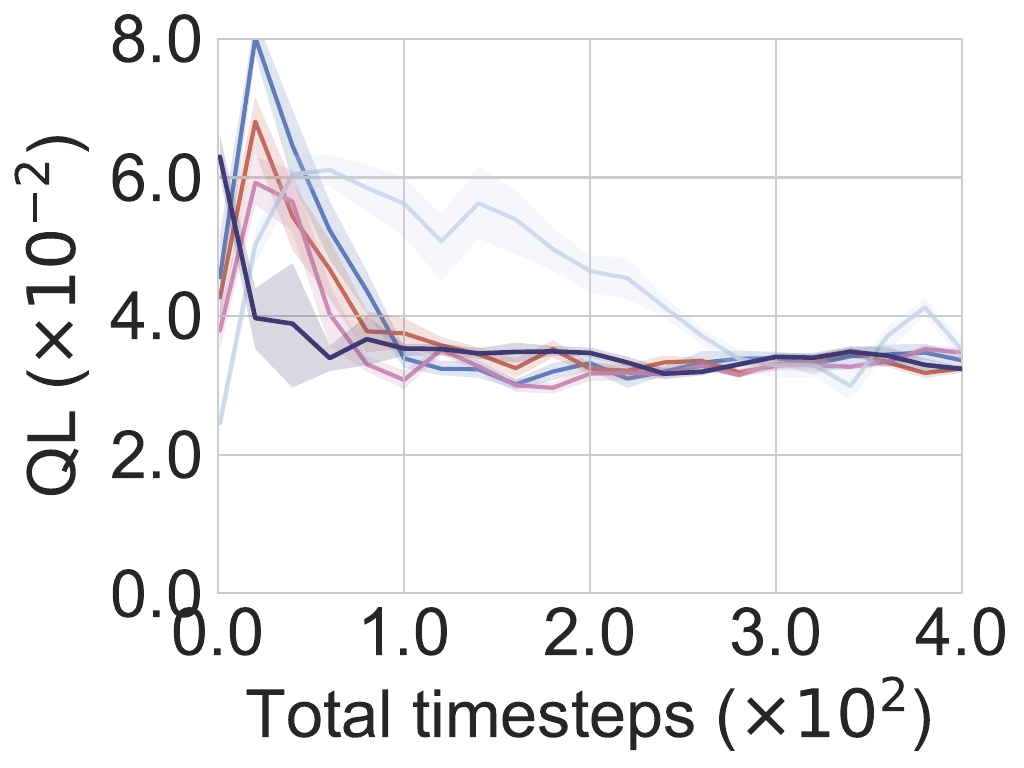} 
    \caption{QL-L1-322}
  \end{subfigure}

  \begin{subfigure}{0.22\textwidth}
    \centering
    \includegraphics[width=0.92\textwidth]{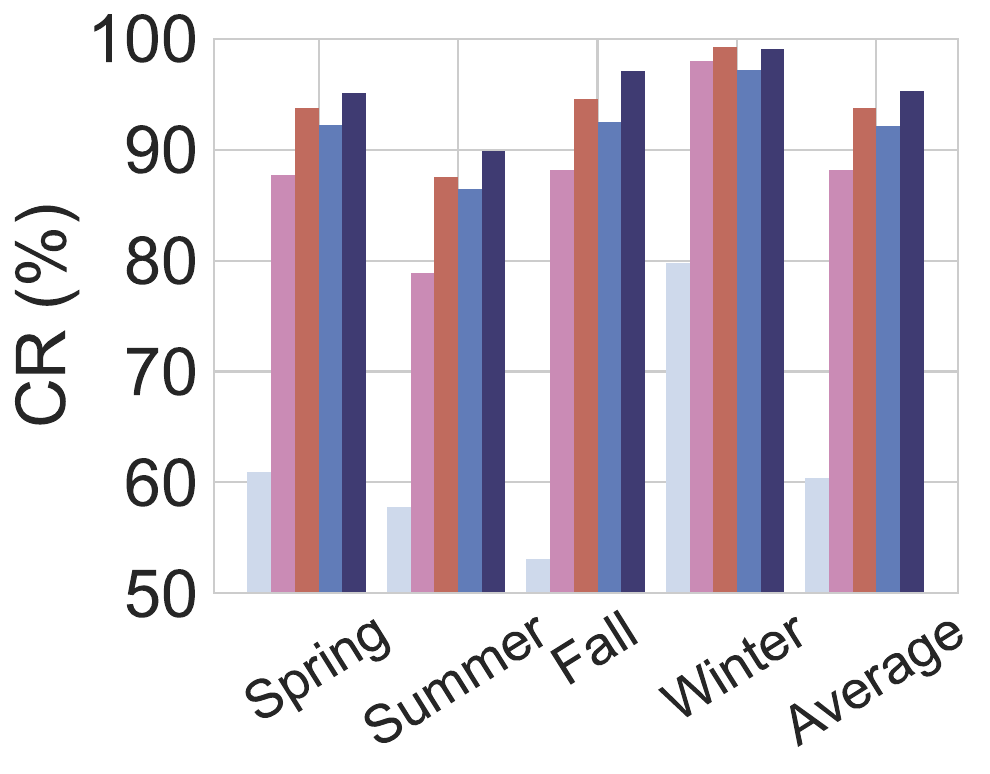}
    \caption{CR-322-Test}
  \end{subfigure}
  \begin{subfigure}{0.22\textwidth}
    \centering
    \includegraphics[width=0.92\textwidth]{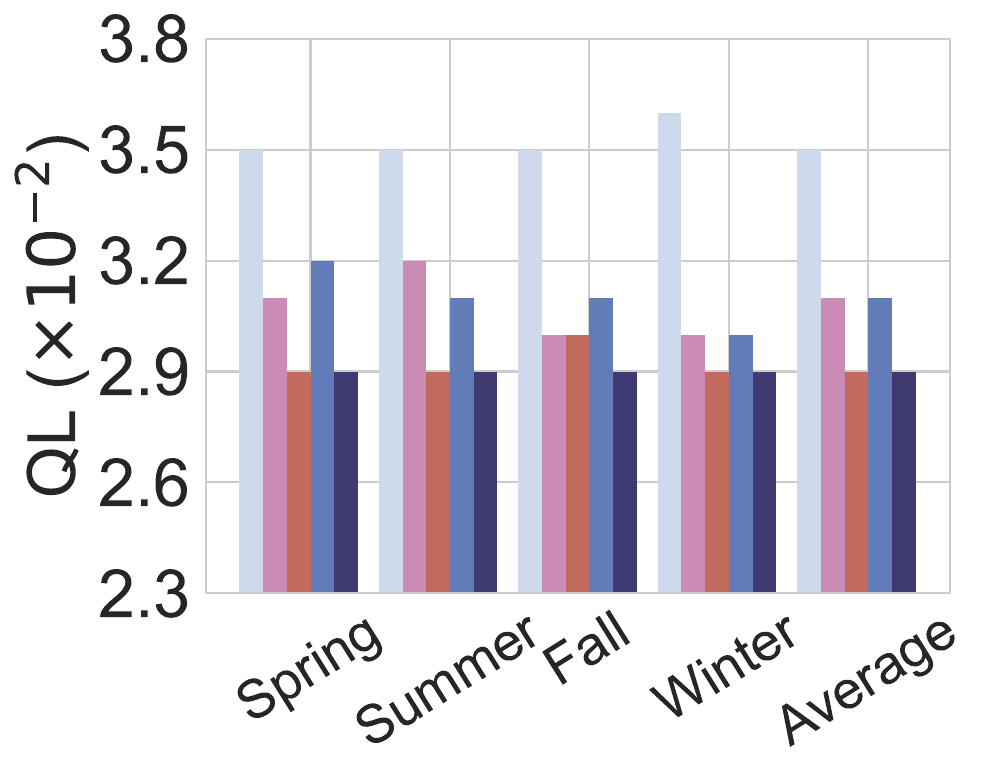} 
    \caption{QL-322-Test}
  \end{subfigure}
  \caption{Training curves and test results of TPA and its ablations on the IEEE 322-bus system. ``S'' denotes the seasonal labels and "M" denotes short-term memory. "P" represents the learnable prototypes. "T-" represents the previous TMAAC method. "TPA" selects MADDPG as the basic algorithm.}
  \label{fig:abs}
\end{figure}

\subsection{Ablation Study}
\label{sec:abl}

To analyze the impact of different components in TPA, we carry out ablation studies to quantitatively evaluate the contribution of different TPA components. The comparison results of various ablations on the IEEE 322-bus system are shown in Figure~\ref{fig:abs}. The training curves present the average metrics.

\textbf{T-MADDPG with S}. To assess the impact of prior temporal knowledge for the TMAAC method, we re-train T-MADDPG on the input short-term data enhanced by the one-hot encoded seasonal feature.
The quantitative results indicate that the re-trained T-MADDPG gets even worse results than the normal performances in Table~\ref{tab:sota}.
The improper processing of the introduced temporal inputs can easily result in training failures, diminishing the ability of encoders to effectively address temporal distribution shifts and hampering their expressiveness.

\textbf{TPA w/o M and TPA w/o S.} To evaluate the effect of short-term memory and the seasonal labels, we implement 2 variant:~1) without short-term memory input, and 2) without the seasonal labels input. The results show that both of them have a positive effect on the final strategy; however, the impact of short-term memory surpasses that of seasonal labels significantly. 
While short-term memory intuitively reflects the current trend, the seasonal labels emphasize capturing broader temporal dependencies. 
In comparison to seasonal labels, short-term memory distinctly captures the trend to facilitate adjustments at each moment, while the sensitivity of seasonal labels to intermittent fluctuations is significantly lower.

\begin{figure}[!t]
  \centering
  
    \begin{subfigure}{0.44\textwidth}
    \centering
    \includegraphics[width=0.6\textwidth] {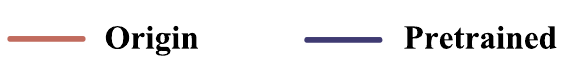}
  \end{subfigure}
  
  \begin{subfigure}{0.22\textwidth}
    \centering
    \includegraphics[width=\textwidth]{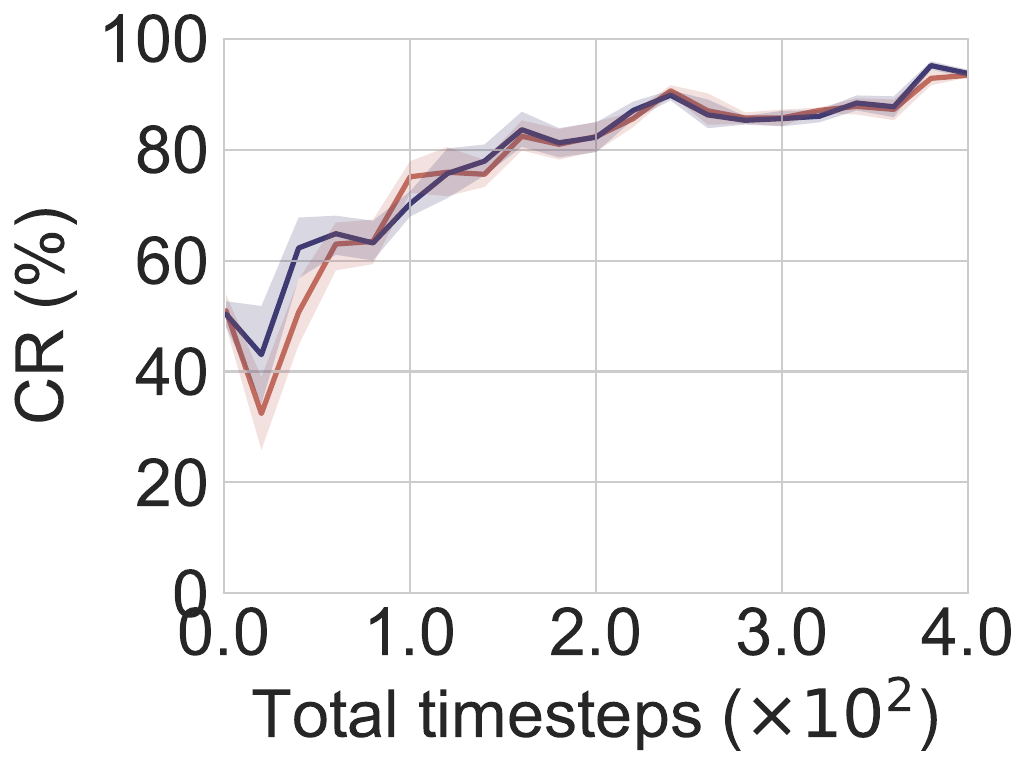}
    \caption{CR-L1-141}
  \end{subfigure}
  \begin{subfigure}{0.22\textwidth}
    \centering
    \includegraphics[width=\textwidth]{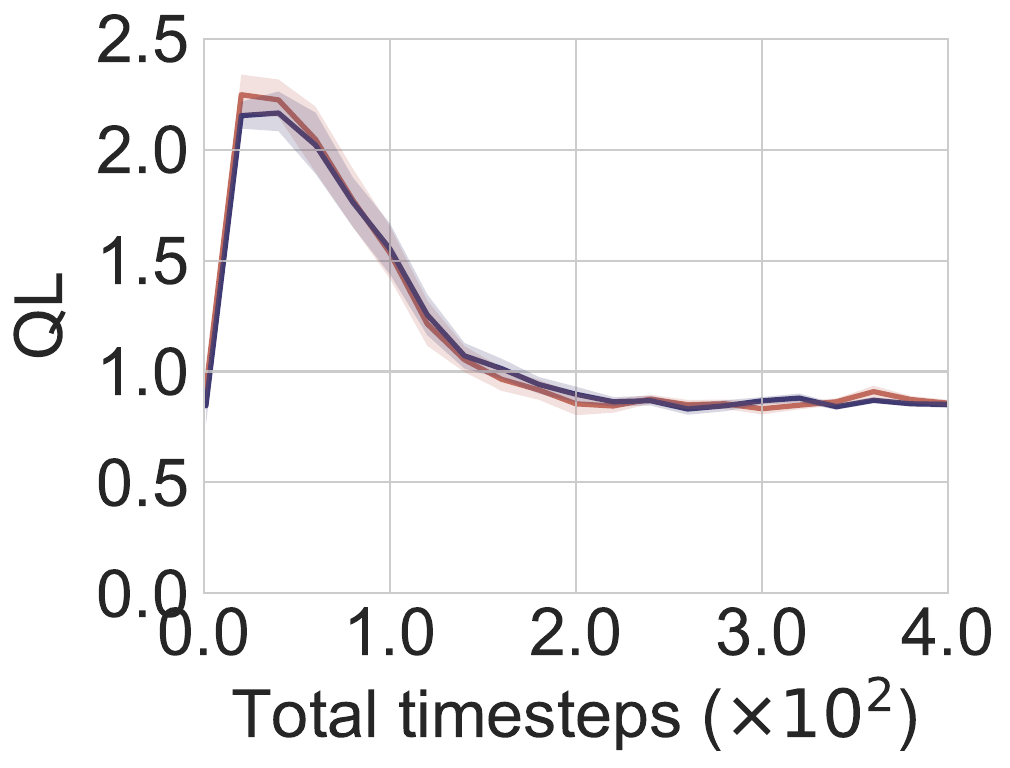} 
    \caption{QL-L1-141}
  \end{subfigure}

    \begin{subfigure}{0.22\textwidth}
    \centering
    \includegraphics[width=0.92\textwidth]{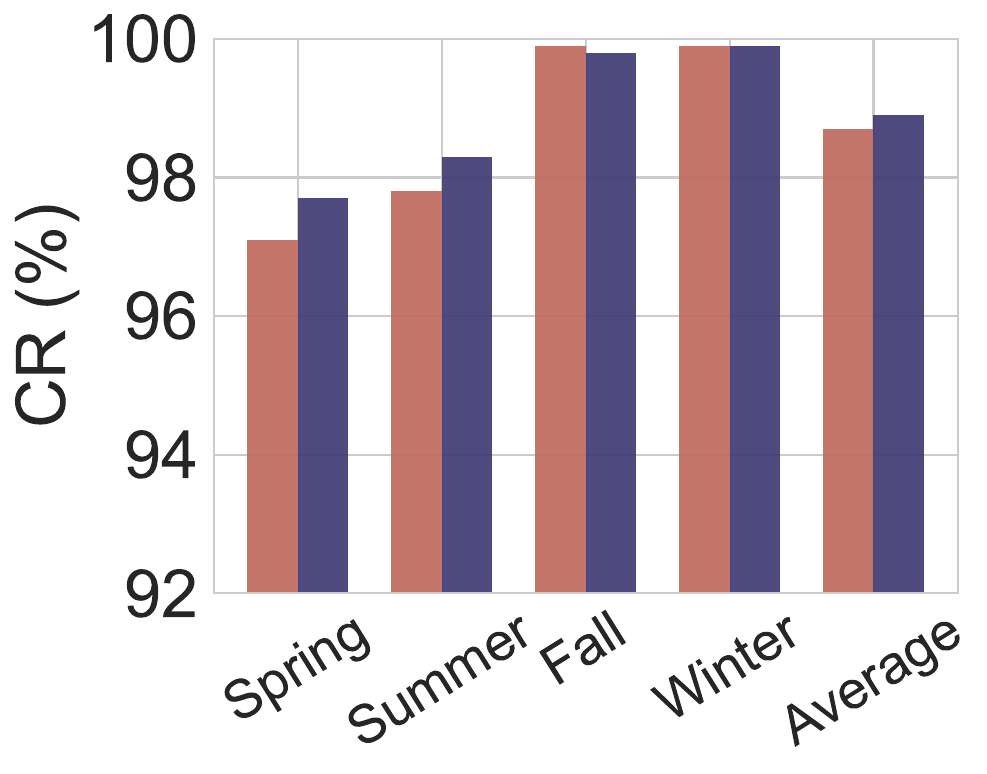}
    \caption{CR-141-Test}
  \end{subfigure}
  \begin{subfigure}{0.22\textwidth}
    \centering
    \includegraphics[width=0.92\textwidth]{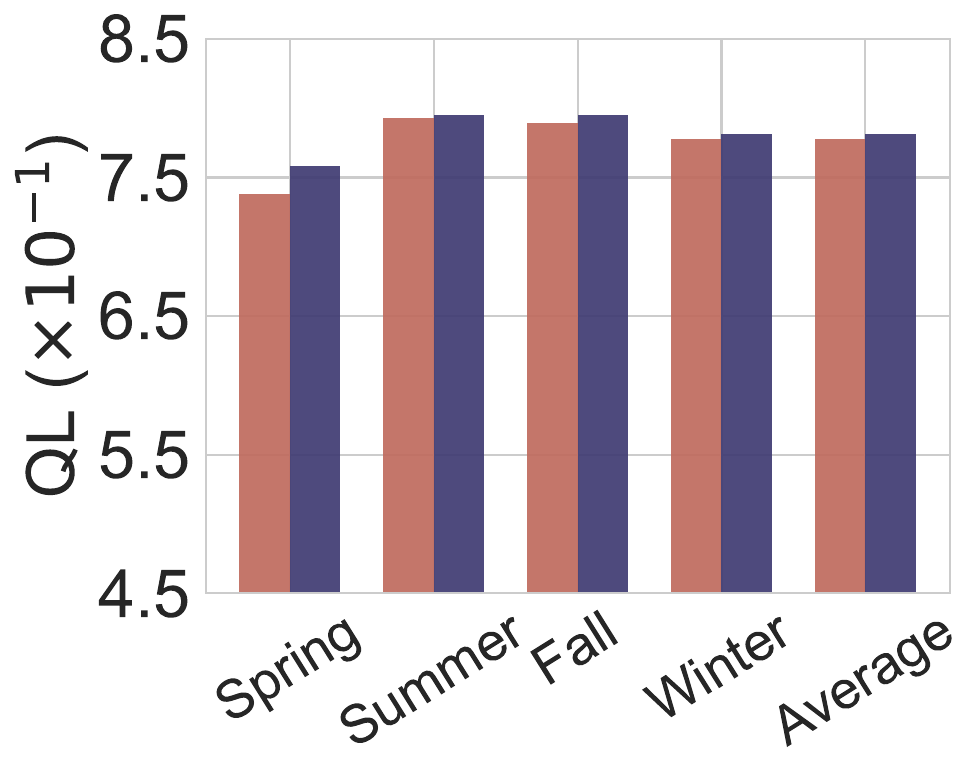} 
    \caption{QL-141-Test}
  \end{subfigure}
\caption{Training curves and test results of transferability experiment on the 141-bus network. "Origin" denotes the normal TPA model. "Pretrained" denotes the TPA model with pre-trained prototypes on the 322-bus network. MADDPG is selected as the basic algorithm.}
\label{fig:transfer}
\end{figure}

\begin{figure*}[!t]
  \centering
    \begin{subfigure}{1.0\textwidth}
    \centering
    \includegraphics[width=0.8\textwidth]{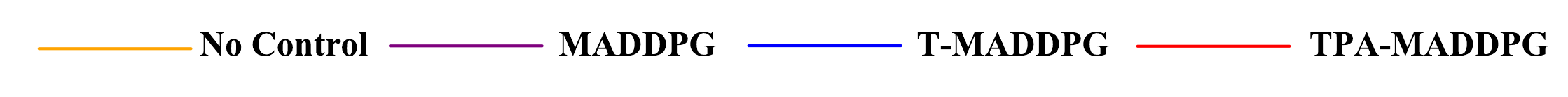}
  \end{subfigure}
  
  \begin{subfigure}{0.23\textwidth}
    \centering
    \includegraphics[width=\textwidth]{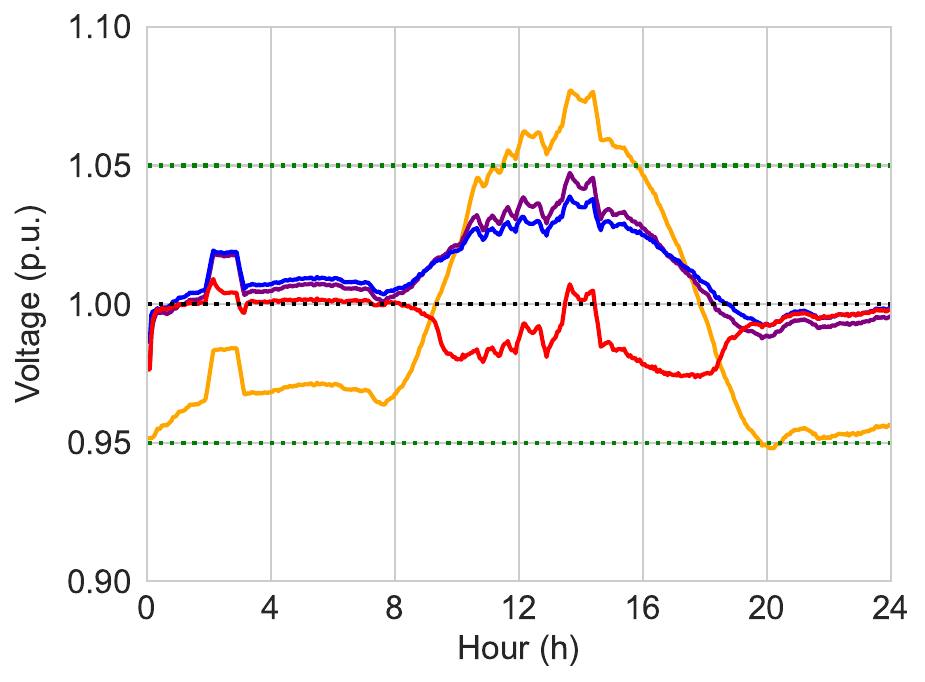}
    \caption{Voltage of bus 254 in spring.}
  \end{subfigure}
  \begin{subfigure}{0.23\textwidth}
    \centering
    \includegraphics[width=\textwidth]{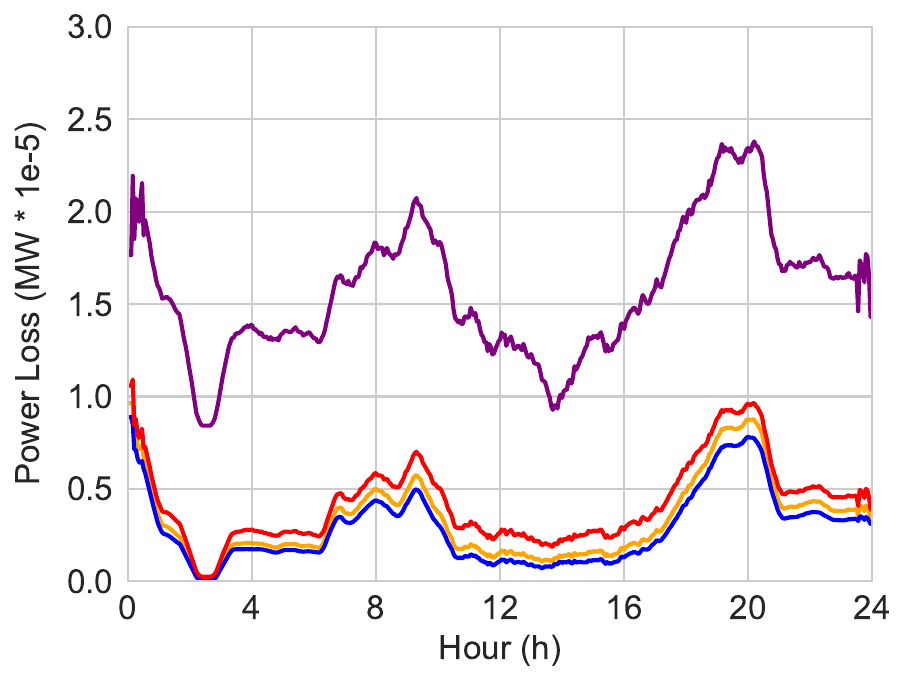} 
    \caption{Loss of bus 254 in spring.}
  \end{subfigure}
  \begin{subfigure}{0.23\textwidth}
    \centering
    \includegraphics[width=\textwidth]{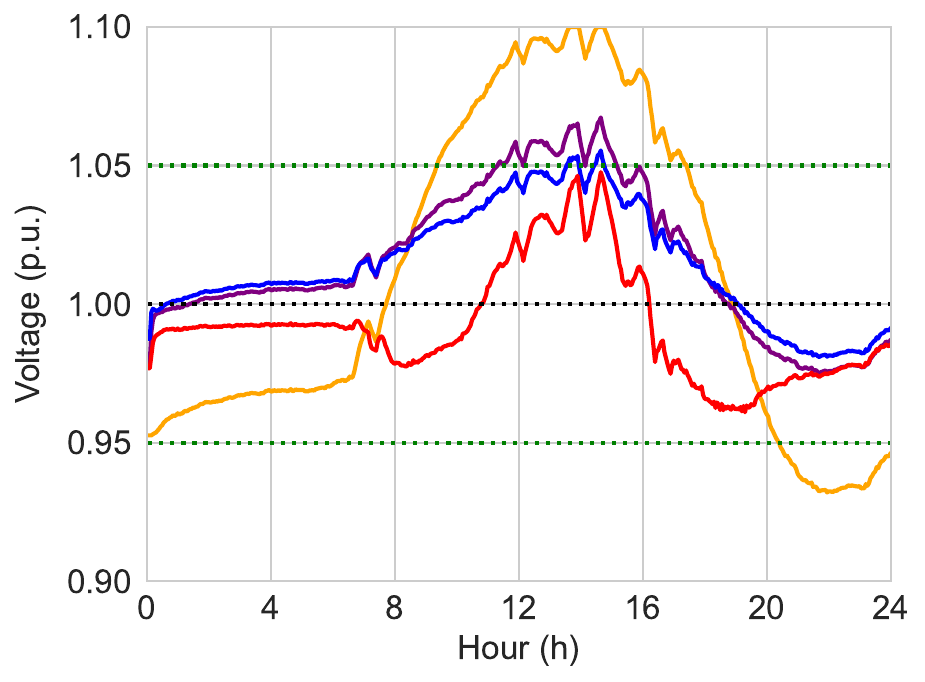} 
    \caption{Voltage of bus 254 in summer.}
  \end{subfigure}
  \begin{subfigure}{0.23\textwidth}
    \centering
    \includegraphics[width=\textwidth]{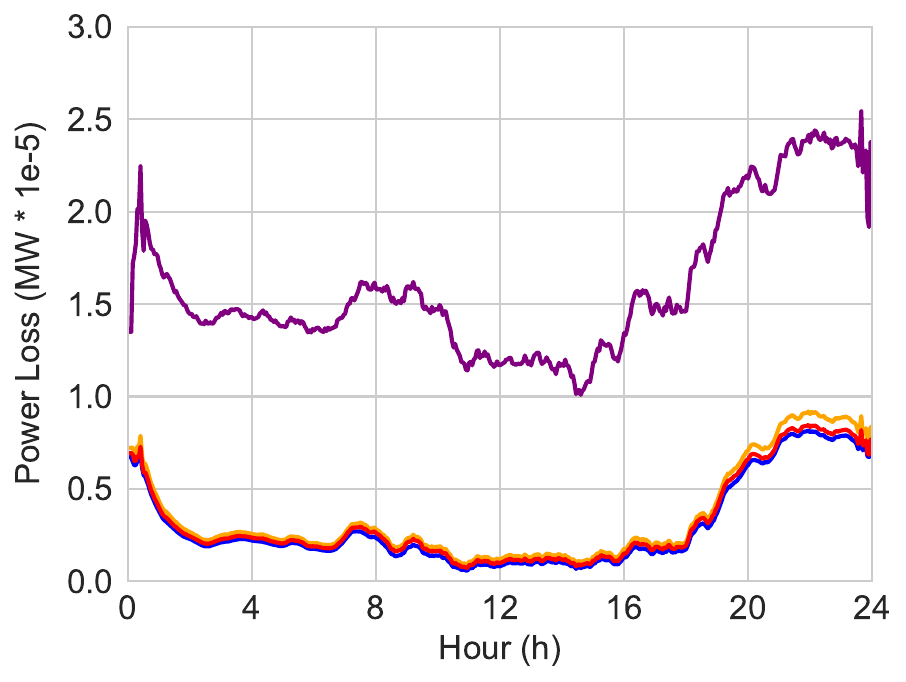} 
    \caption{Loss of bus 254 in summer.}
  \end{subfigure}
    \begin{subfigure}{0.23\textwidth}
    \centering
    \includegraphics[width=\textwidth]{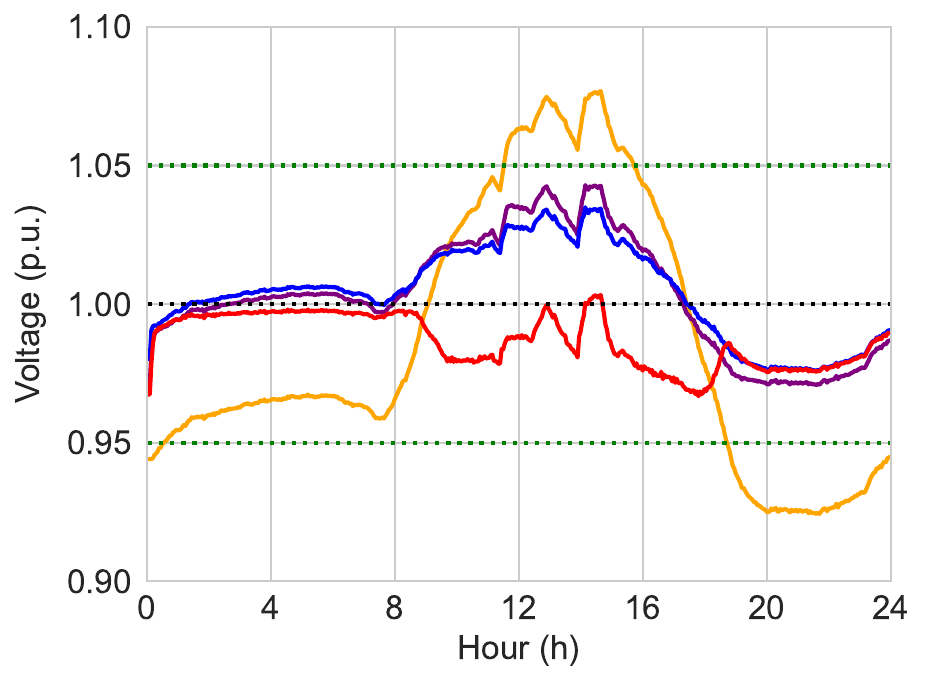} 
    \caption{Voltage of bus 254 in fall.}
  \end{subfigure}
  \begin{subfigure}{0.23\textwidth}
    \centering
    \includegraphics[width=\textwidth]{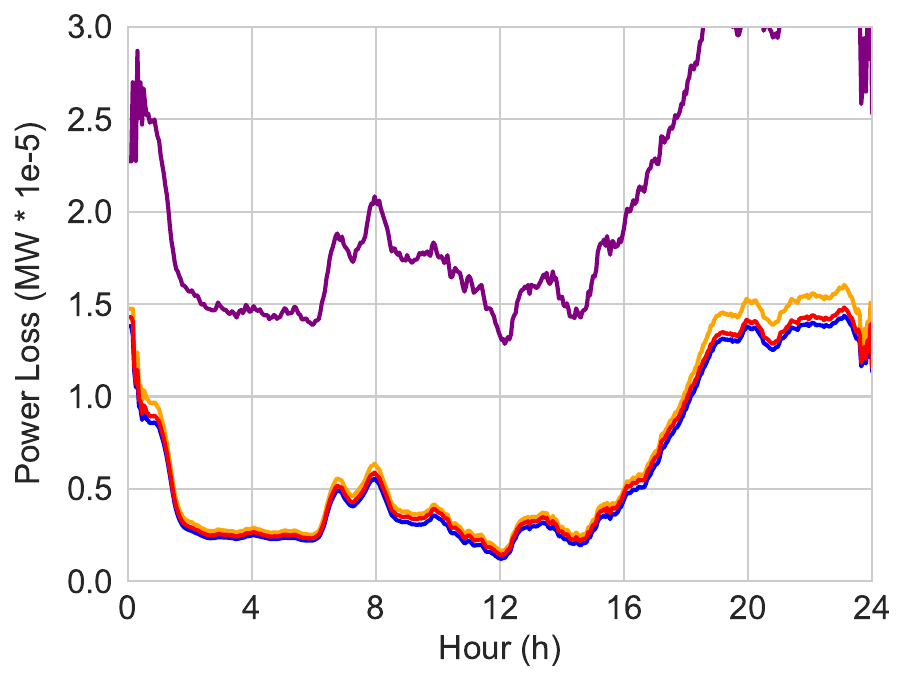} 
    \caption{Loss of bus 254 in fall.}
  \end{subfigure}
      \begin{subfigure}{0.23\textwidth}
    \centering
    \includegraphics[width=\textwidth]{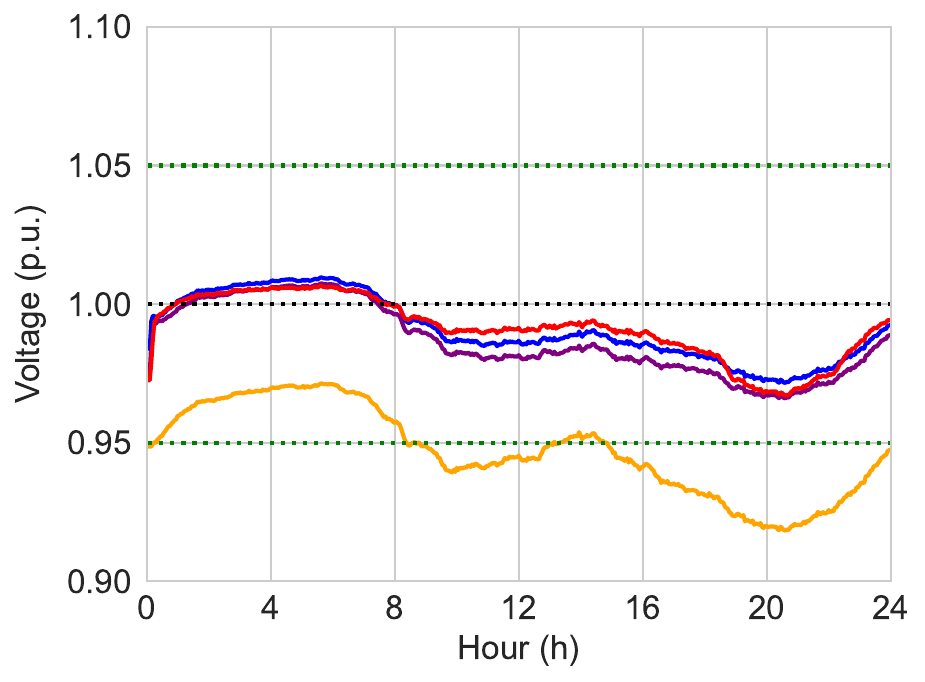} 
    \caption{Voltage of bus 254 in winter.}
  \end{subfigure}
  \begin{subfigure}{0.23\textwidth}
    \centering
    \includegraphics[width=\textwidth]{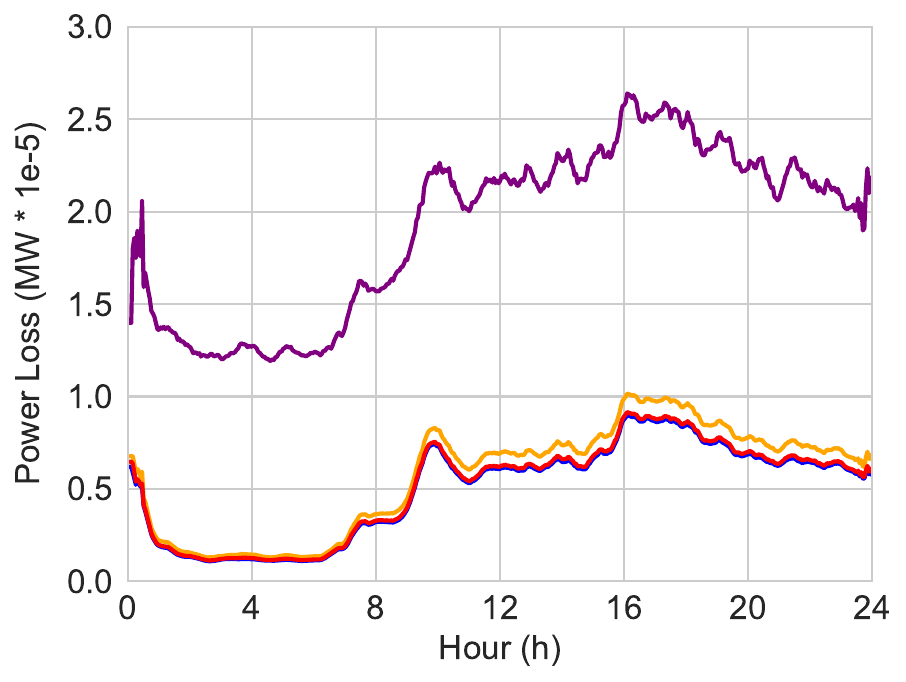} 
    \caption{Loss of bus 254 in winter.}
  \end{subfigure}
  \caption{Compare TPA with other methods on a bus during one day for the 322-bus network. The green and black dotted lines in~(a) represent the safety voltage range and ideal voltage respectively.}
  \label{fig:vis}
\end{figure*}

\textbf{TPA w/o P.} To examine the effectiveness of the temporal prototype, we remove the temporal prototype to predict the action based on encoded features. 
The results show the significant impact of temporal prototypes on the stability of the learning process.
Figure~\ref{fig:abs} illustrates that "TPA w/o P" only obtained suboptimal results.
Although the former encoder pays attention to multi-scale dependencies, the agent performance deteriorates and wavers around optimal decisions under seasonal fluctuations, due to the lack of season climates introduced by temporal prototypes.

\subsection{Transferability Analysis}
\label{sec:trans}

To further evaluate the transferability of the generated temporal prototypes, we evaluate the TPA performance on the 141-bus network with fixed temporal prototypes trained under the 322-bus network.
Concretely, we fully trained a TPA model on the 322-bus network to acquire the pre-trained temporal prototypes. Subsequently, in the new 141-bus environment, we re-trained a new TPA model while initializing it with these pre-trained temporal prototypes and keeping them fixed during the training process.

The results compared to the original TPA methods are presented in Figure~\ref{fig:transfer}. 
The training curves and testing results indicate that TPA with pre-trained prototypes from other networks performs equally well on the target network. In the spring, summer, and the final average CR metric, TPA with pre-trained prototypes can even surpass the original TPA method. 
TPA with pre-trained prototypes also has a higher CR at the beginning stages of training.
The improved performance could be attributed to the precise guidance offered by pre-trained prototypes, assisting the former encoder module in learning with temporal dependencies.
While conventional MARL methods dynamically adjust the model structure based on the varying scales of the power networks, our prototypes provide global temporal patterns that are independent of network topology.

\subsection{Visualization Analysis}
\label{sec:va}

To further explain the learned time-adaptive strategies from TPA, we conduct a qualitative analysis. The voltage and power loss of a single bus over one day across different seasons under the 322-bus network are visualized in Figure~\ref{fig:vis}. We compared our TPA-MADDPG method with no control, common MADDPG, and the previous state-of-the-art T-MADDPG.
The visual results show that the proposed method successfully learns time-adaptive policies for different seasons, which significantly stabilizes the voltage.
Specifically, in extreme cases such as the noon in summer, a large penetration of PVs leads to reverse current flow that would increase $v_i$ out of the nominal range. Benefiting from multi-scale temporal offset adaptation, the TPA method can effectively avoid risks at such moments and ensure safety by controlling the overall voltage within a lower range during summer.

\section{Conclusions}

In this paper, we explore the challenge of learning long-term time-adaptive policy under short-term training trajectories. Unlike the previous methods within the confines of a day cycle, we propose a temporal prototype-aware~(TPA) method to effectively control even on month and year cycles. The proposed method can not only learn underlying temporal dependencies by the multi-scale dynamic encoder but also dynamically adapt to the evolving season climates by matching the temporal prototypes. Experimental results show that TPA outperforms other methods in both controllable rate and power generation loss, especially in the long-term operation cycles. Moreover, TPA is readily applicable to various MARL methods and shows the transferability of prototypes across different PDN sizes.
In our future work, we will enhance the interpretability of our model to ensure more reliable practical applications.

\begin{acks}
This work was supported in part by the Joint Funds of the Zhejiang Provincial Natural Science Foundation of China under Grant LHZSD24F020001, in part by the Zhejiang Province ``LingYan" Research and Development Plan Project under Grant 2024C01114, and in part by the Zhejiang Province High-Level Talents Special Support Program ``Leading Talent of Technological Innovation of Ten-Thousands Talents Program" under Grant 2022R52046.
\end{acks}

\bibliographystyle{ACM-Reference-Format}
\balance
\bibliography{sample-base}

\appendix

\newpage

\section{More Implementation Details}
\label{sec:apx_da}

\subsection{Datasets}

The 141-bus network is divided into 9 zones and contains 84 loads and 22 PVs. The 322-bus network is divided into 22 zones and contains 337 loads and 38 PVs. Following the previous work~\cite{MAPDN}, we set the range of actions for each scenario, $[ -0.6, 0.6]$ for 141-bus network, and $[ -0.8, 0.8]$ for 322-bus network . To tune the trade-off between CR and QL, it is suggested that $\alpha$ of reward function in Eq.~(\ref{eq:rf}) is set to 0.1 in the 322-bus network and 0.01 in the 141-bus network~\cite{MAPDN}. Each time step represents a duration of 3 minutes. So every training episode lasts for 240 time steps and every singular diurnal cycle testing lasts for 480 time steps.

\begin{figure}[!h]
  \centering
  \begin{subfigure}{0.15\textwidth}
    \centering
    \includegraphics[width=\textwidth]{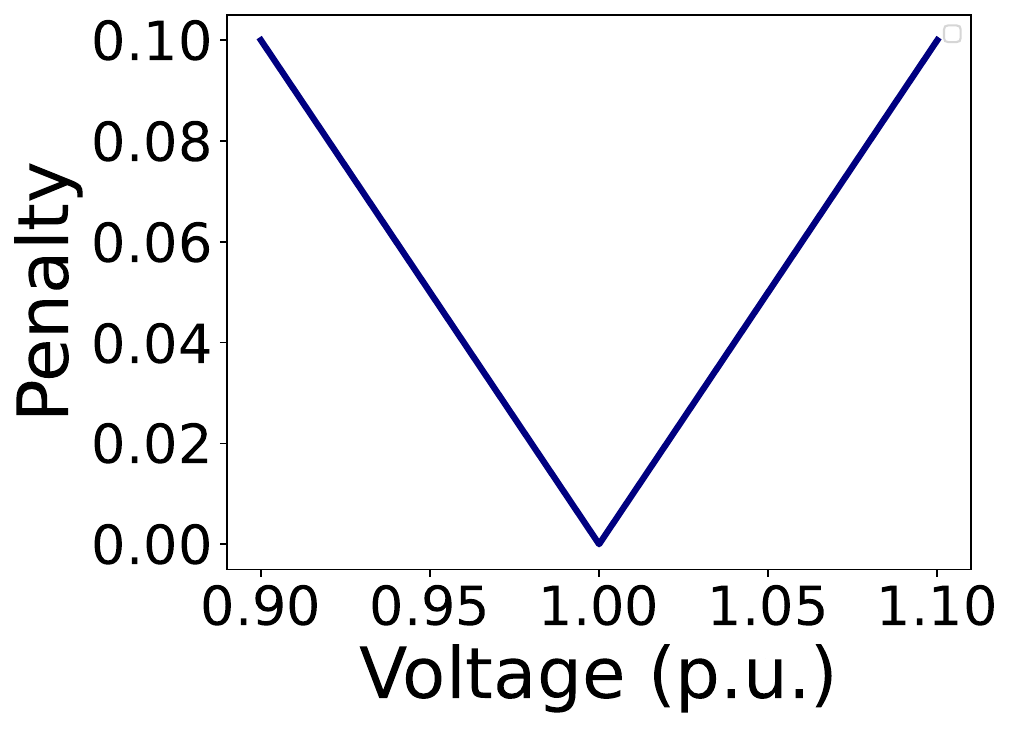}
    \caption{L1-Shape}
  \end{subfigure}
    \begin{subfigure}{0.15\textwidth}
    \centering
    \includegraphics[width=\textwidth]{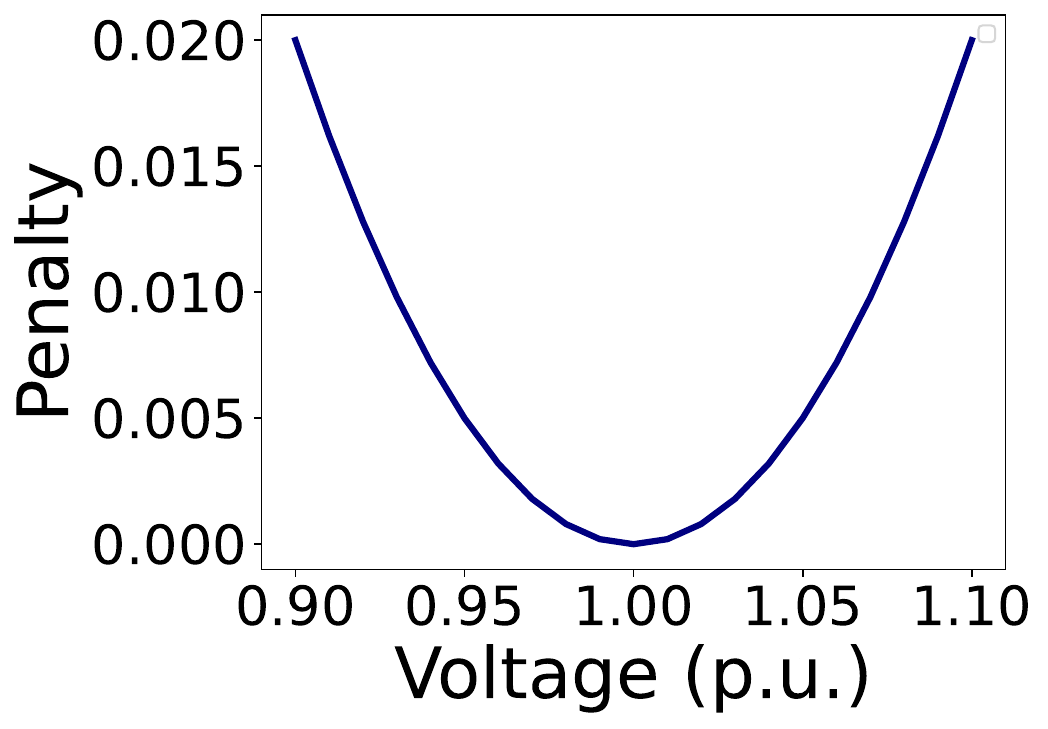}
    \caption{L2-Shape}
  \end{subfigure}
    \begin{subfigure}{0.15\textwidth}
    \centering
    \includegraphics[width=\textwidth]{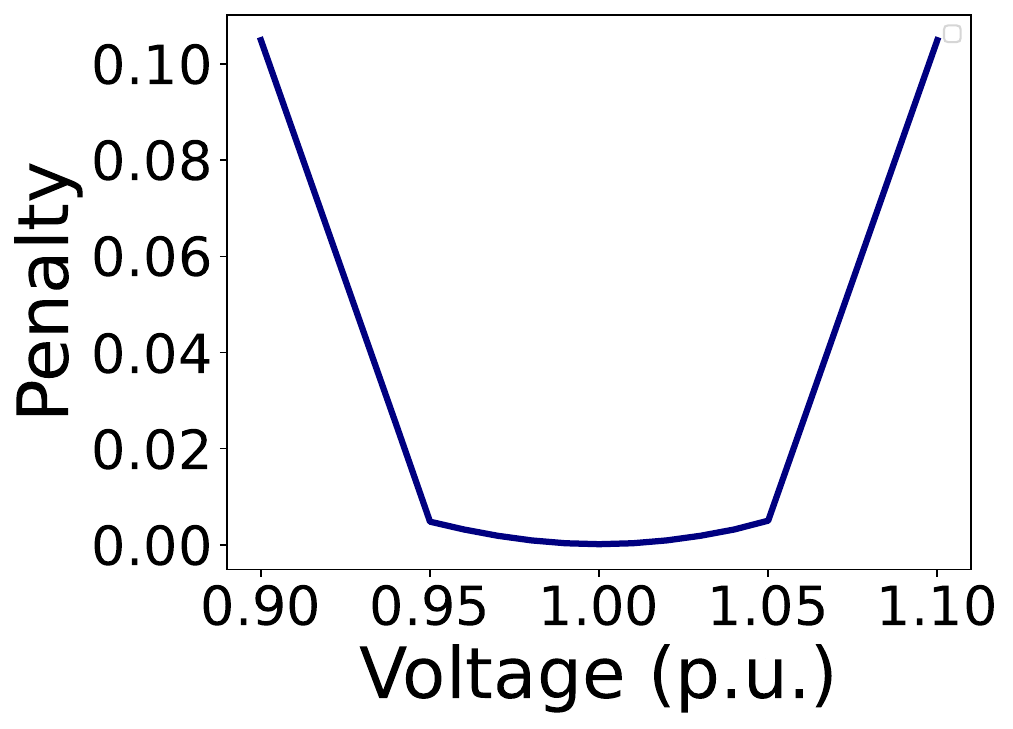}
    \caption{BOWL-Shape}
  \end{subfigure}
\caption{Three types of voltage barrier function.}
\label{fig:barrier}
\end{figure}

There are three types of different voltage barrier functions proposed in the MAPDN environment~\cite{MAPDN}, as shown in Figure~\ref{fig:barrier}. The voltage constraint is difficult to handle in MARL, so we need to use a barrier function to represent the constraint. Each barrier function has a different objective, for example, L1-Shape will severely punish fluctuating voltage, but fluctuations within a safe range are acceptable, which can easily lead to voltage waste. L2-Shape is the opposite, which can easily lead to exceeding the safety threshold. Bowl-Shape has made a compromise by combining the characteristics of both. We conduct the main experiments with different voltage barrier functions. In most experiments, L1-Shape can provide the best performance.

\subsection{Training and Testing.} During training, each experiment is run with 5 random seeds. Each experiment is evaluated on the validation dataset~(1 day per month) every 20 episodes and the evaluation results during training are given by the 50\% ci shading. For singular diurnal cycle testing, we use the test dataset provided by~\cite{TMAAC} which picks 10 days per month and has 120 days in total. For the longer testing cycles, we divide the data for the entire 2014 year into continuous operation times of one month and one year to test existing strategies. During testing, we follow the previous work to choose the best voltage barrier function for every method.

\subsection{Implementation Details.} Following previous works~\cite{MAPDN,TMAAC}, we train our model in a total of 400 epochs. We also use the same optimizer, learning rate, and update frequency to optimize our model. The batch size is 32 during training and the hidden dimension is 64 for all layers~\cite{MAPDN,TMAAC}. Each transformer-based encoder has a 3-layer transformer~\cite{TMAAC}. All experiments use the same hardware and software configuration. 

\section{Additional Experiments}
\subsection{Experiment Results on year-times cycles}
\begin{table}[!h]
\caption{Test results on the IEEE 322-bus system under longer year-times cycles. Bold denotes the best results.}
\resizebox{1.0\columnwidth}{!}{
  \begin{tabular}{@{}ll|cccccc@{}}
  \toprule
  \multicolumn{2}{c|}{\multirow{2}{*}{Method}} & \multicolumn{2}{c}{1-Year}                          & \multicolumn{2}{c}{2-Year}                          & \multicolumn{2}{c}{3-Year}     \\ \cmidrule(l){3-8} 
  \multicolumn{2}{c|}{}                        & CR            & \multicolumn{1}{c|}{QL}             & CR            & \multicolumn{1}{c|}{QL}             & CR            & QL             \\ \midrule
  \multicolumn{2}{l|}{322-MADDPG}              & 85.5          & \multicolumn{1}{c|}{0.032}          & 86.2          & \multicolumn{1}{c|}{\textbf{0.029}} & 87.6          & 0.030          \\
  \multicolumn{2}{l|}{322-T-MADDPG}            & 88.2          & \multicolumn{1}{c|}{0.031}          & 88.9          & \multicolumn{1}{c|}{0.032}          & 89.1          & 0.031          \\ \midrule
  \multicolumn{2}{l|}{\cellcolor{gray!13}322-TPA-MADDPG}          & \textbf{92.2} & \multicolumn{1}{c|}{\textbf{0.030}} & \textbf{93.9} & \multicolumn{1}{c|}{\textbf{0.029}}          & \textbf{94.3} & \textbf{0.029} \\ \bottomrule
\end{tabular}
}
\label{tab:yeartimes}
\end{table}

As shown in Table~\ref{tab:yeartimes}, we also conduct experiments on the longer year-times cycles.
Extending the test cycle on year-times increases the overall performance because the stable and easy periods of the year are significantly longer than the unstable and extreme periods. It can still illustrate the inadequacies of current AVC methods in addressing long-term operational variability, while our TPA model has achieved the best performance in all year-times cycle tests. The advantage becomes more apparent with the increase in test cycles. 

\subsection{Same season divisions as TMAAC}

\begin{table}[!h]
\caption{Test results with the same season division as TMAAC. Bold denotes the best results. ± corresponds to the standard deviation of testing episodes.}
\resizebox{1.0\columnwidth}{!}{
  \begin{tabular}{@{}ll|cccccccccc@{}}
  \toprule
  \multicolumn{2}{c|}{\multirow{2}{*}{Methods}} & \multicolumn{2}{c}{Spring}                      & \multicolumn{2}{c}{Summer}                      & \multicolumn{2}{c}{Fall}                        & \multicolumn{2}{c}{Winter}                     & \multicolumn{2}{c}{All}        \\ \cmidrule(l){3-12} 
  \multicolumn{2}{c|}{}                         & CR        & \multicolumn{1}{c|}{QL}             & CR        & \multicolumn{1}{c|}{QL}             & CR        & \multicolumn{1}{c|}{QL}             & CR       & \multicolumn{1}{c|}{QL}             & CR            & QL             \\ \midrule
  \multicolumn{2}{l|}{322-MATD3}                & 80.7{\scalebox{0.8}{$\pm$20.2}} & \multicolumn{1}{c|}{0.039}          & 77.3{\scalebox{0.8}{$\pm$18.2}} & \multicolumn{1}{c|}{0.039}          & 93.5{\scalebox{0.8}{$\pm$13.4}} & \multicolumn{1}{c|}{0.038}          & 98.3{\scalebox{0.8}{$\pm$}}3.9 & \multicolumn{1}{c|}{0.038}          & 87.4          & 0.038          \\
  \multicolumn{2}{l|}{322-MADDPG}               & 80.1{\scalebox{0.8}{$\pm$18.9}} & \multicolumn{1}{c|}{0.033}          & 74.6{\scalebox{0.8}{$\pm$19.8}} & \multicolumn{1}{c|}{0.033}          & 92.7{\scalebox{0.8}{$\pm$14.0}} & \multicolumn{1}{c|}{0.032}          & 96.7{\scalebox{0.8}{$\pm$}}7.6 & \multicolumn{1}{c|}{0.031}          & 86.0          & 0.032          \\
  \multicolumn{2}{l|}{322-T-MATD3}              & 90.8{\scalebox{0.8}{$\pm$17.5}} & \multicolumn{1}{c|}{\textbf{0.028}} & 88.8{\scalebox{0.8}{$\pm$20.4}} & \multicolumn{1}{c|}{0.032}          & 97.4{\scalebox{0.8}{$\pm$8.1}}  & \multicolumn{1}{c|}{\textbf{0.029}} & 97.2{\scalebox{0.8}{$\pm$}}8.6 & \multicolumn{1}{c|}{\textbf{0.028}} & 93.5          & \textbf{0.029} \\
  \multicolumn{2}{l|}{322-T-MADDPG}             & 87.8{\scalebox{0.8}{$\pm$19.8}} & \multicolumn{1}{c|}{0.032}          & 83.3{\scalebox{0.8}{$\pm$20.7}} & \multicolumn{1}{c|}{0.033}          & 96.2{\scalebox{0.8}{$\pm$8.7}}  & \multicolumn{1}{c|}{0.031}          & 97.1{\scalebox{0.8}{$\pm$}}8.5 & \multicolumn{1}{c|}{0.030}          & 91.1          & 0.031          \\ \midrule
  \multicolumn{2}{l|}{\cellcolor{gray!13}322-TPA-MATD3}            & 89.9{\scalebox{0.8}{$\pm$18.7}} & \multicolumn{1}{c|}{\textbf{0.028}} & 86.7{\scalebox{0.8}{$\pm$21.2}} & \multicolumn{1}{c|}{0.032}          & 96.2{\scalebox{0.8}{$\pm$9.1}}  & \multicolumn{1}{c|}{\textbf{0.029}} & 97.2{\scalebox{0.8}{$\pm$}}5.7 & \multicolumn{1}{c|}{0.029}          & 92.5          & 0.030          \\
  \multicolumn{2}{l|}{\cellcolor{gray!13}322-TPA-MADDPG}           & \textbf{92.7}{\scalebox{0.8}{$\pm$17.1}} & \multicolumn{1}{c|}{\textbf{0.028}} & \textbf{91.1}{\scalebox{0.8}{$\pm$20.3}} & \multicolumn{1}{c|}{\textbf{0.030}} & \textbf{98.4}{\scalebox{0.8}{$\pm$5.7}}  & \multicolumn{1}{c|}{\textbf{0.029}} & \textbf{99.0}{\scalebox{0.8}{$\pm$}}3.3 & \multicolumn{1}{c|}{\textbf{0.028}} & \textbf{95.3} & \textbf{0.029} \\ \bottomrule
\end{tabular}
}
\label{tab:monthdiv}
\end{table}

We have conducted experiments on the same month divisions for each season as TMAAC, as shown in Table~\ref{tab:monthdiv}. In our original setting, we used different month divisions compared to TMAAC. For example, we divide spring into 2, 3, and 4. While they divide spring into 3, 4, and 5. After unifying the settings, their original 322-T-MATD3 achieved 90.6\% in spring, whereas our 322-T-MATD3 achieved 93.5\%. The new results are also more similar to the values they reported in TMAAC~\citep{TMAAC}.

\subsection{Experiments on 33-bus network}

\begin{table}[!h]
\caption{Test results of algorithms in 33-bus system. Bold denotes the best results. ± corresponds to the standard deviation of testing episodes.}
\resizebox{1.0\columnwidth}{!}{
    \begin{tabular}{@{}ll|cccccccccc@{}}
    \toprule
    \multicolumn{2}{c|}{\multirow{2}{*}{Methods}} & \multicolumn{2}{c}{Spring}                                                       & \multicolumn{2}{c}{Summer}                                                       & \multicolumn{2}{c}{Fall}                                                         & \multicolumn{2}{c}{Winter}                                                       & \multicolumn{2}{c}{Average} \\ \cmidrule(l){3-12} 
    \multicolumn{2}{c|}{}                         & CR                                                   & \multicolumn{1}{c|}{QL}   & CR                                                   & \multicolumn{1}{c|}{QL}   & CR                                                   & \multicolumn{1}{c|}{QL}   & CR                                                   & \multicolumn{1}{c|}{QL}   & CR           & QL           \\ \midrule
    \multicolumn{2}{l|}{33-MATD3}                 & 99.5{\scalebox{0.8}{$\pm$0.02}} & \multicolumn{1}{c|}{0.53} & 98.6{\scalebox{0.8}{$\pm$0.05}} & \multicolumn{1}{c|}{0.49} & 99.9{\scalebox{0.8}{$\pm$0.01}} & \multicolumn{1}{c|}{0.53} & \textbf{100.0}{\scalebox{0.8}{$\pm$0.0}}   & \multicolumn{1}{c|}{0.59} & 99.5         & 0.54        \\
    \multicolumn{2}{l|}{33-MADDPG}                & 99.0{\scalebox{0.8}{$\pm$0.04}}   & \multicolumn{1}{c|}{0.36} & 97.8{\scalebox{0.8}{$\pm$0.08}} & \multicolumn{1}{c|}{0.32} & 99.9{\scalebox{0.8}{$\pm$0.01}} & \multicolumn{1}{c|}{0.36} & 99.9{\scalebox{0.8}{$\pm$0.01}} & \multicolumn{1}{c|}{0.41} & 99.2         & 0.36        \\
    \multicolumn{2}{l|}{33-T-MADDPG}              & 98.9{\scalebox{0.8}{$\pm$0.04}} & \multicolumn{1}{c|}{0.41} & 97.6{\scalebox{0.8}{$\pm$0.09}} & \multicolumn{1}{c|}{0.37} & 99.6{\scalebox{0.8}{$\pm$0.01}} & \multicolumn{1}{c|}{0.41} & \textbf{100.0}{\scalebox{0.8}{$\pm$0.0}}   & \multicolumn{1}{c|}{0.48} & 99.0          & 0.42        \\
    \multicolumn{2}{l|}{33-T-MATD3}               & \textbf{99.9}{\scalebox{0.8}{$\pm$0.01}} & \multicolumn{1}{c|}{0.34} & \textbf{99.9}{\scalebox{0.8}{$\pm$0.01}} & \multicolumn{1}{c|}{0.35} & \textbf{100.0}{\scalebox{0.8}{$\pm$0.0}}   & \multicolumn{1}{c|}{0.34} & 99.9{\scalebox{0.8}{$\pm$0.01}} & \multicolumn{1}{c|}{0.35} & \textbf{99.9}         & 0.35        \\ \midrule
    \multicolumn{2}{l|}{\cellcolor{gray!13}33-TPA-MATD3}             & 99.5{\scalebox{0.8}{$\pm$0.02}} & \multicolumn{1}{c|}{\textbf{0.25}} & 98.9{\scalebox{0.8}{$\pm$0.03}} & \multicolumn{1}{c|}{\textbf{0.26}} & 99.9{\scalebox{0.8}{$\pm$0.01}} & \multicolumn{1}{c|}{\textbf{0.25}} & \textbf{100.0}{\scalebox{0.8}{$\pm$0.0}}   & \multicolumn{1}{c|}{\textbf{0.25}} & 99.6         & \textbf{0.25}       \\
    \multicolumn{2}{l|}{\cellcolor{gray!13}33-TPA-MADDPG}            & \textbf{99.9}{\scalebox{0.8}{$\pm$0.01}} & \multicolumn{1}{c|}{0.26} & \textbf{99.9}{\scalebox{0.8}{$\pm$0.01}} & \multicolumn{1}{c|}{\textbf{0.26}} & \textbf{100.0}{\scalebox{0.8}{$\pm$0.0}}   & \multicolumn{1}{c|}{0.26} & \textbf{100.0}{\scalebox{0.8}{$\pm$0.0}}  & \multicolumn{1}{c|}{0.26} & \textbf{99.9}        & 0.26         \\ \bottomrule
\end{tabular}
}
\label{tab:33bus}
\end{table}

We compare the performances of our TPA and other baselines under the singular diurnal cycle on the 33-bus network. Test performances are shown in Table~\ref{tab:33bus}.
In smaller networks, the control rates~(CR) between all methods are maintained at a high level, and power generation loss~(QL) becomes the key to evaluating model performance. This trend is very similar to the trend on the 141-bus system. Our TPA method still achieved the optimal CR and QL. 

\subsection{Resources Consumption}

We conducted experiments on a basic MADDPG under the 33-bus network to investigate the higher computational resources brought by longer trajectories. 

\begin{table}[!h]
\caption{Longer trajectories under 33-bus network.}
\resizebox{1.0\columnwidth}{!}{
  \begin{tabular}{@{}lcccc@{}}
  \toprule
  Trajectory Length  & half-day(240) & one day(480) & one month(14400) & one year(175200) \\ \midrule
  Get one trajectory & 0.89s         & 1.53s        & 2.05s            & 3.42s            \\
  Train one epoch    & 24.5s         & 55.4s        & 1170.9s          & 16534.8s         \\ \bottomrule
\end{tabular}
}
\label{tab:computersource}
\end{table}

As shown in Table~\ref{tab:computersource}, we extend the training trajectory of a half day to one day, one month and even one year. 
Previous works and our TPA set 240 timesteps as the default training segment, which is appropriate for the total data capacity of three years and fast model interaction time. The extension of trajectory length has minimal impact on getting trajectories, but it has very serious requirements for model training resources. If we want to incorporate seasonal climates into training segments, the training time will increase by thousands of times. Therefore, simply lengthening the training trajectory is extremely wasteful of resources.
It is also should be noted that our TPA method and other methods did not utilize long trajectories for training in our original manuscript. 

\subsection{Randomly Initialized Prototypes}

\begin{figure}[H]
  \begin{subfigure}{0.4\textwidth}
    \centering
    \includegraphics[width=\textwidth]{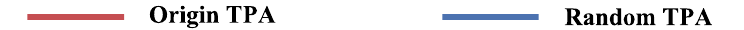}
  \end{subfigure}
  
  \begin{subfigure}{0.22\textwidth}
    \centering
    \includegraphics[width=\textwidth]{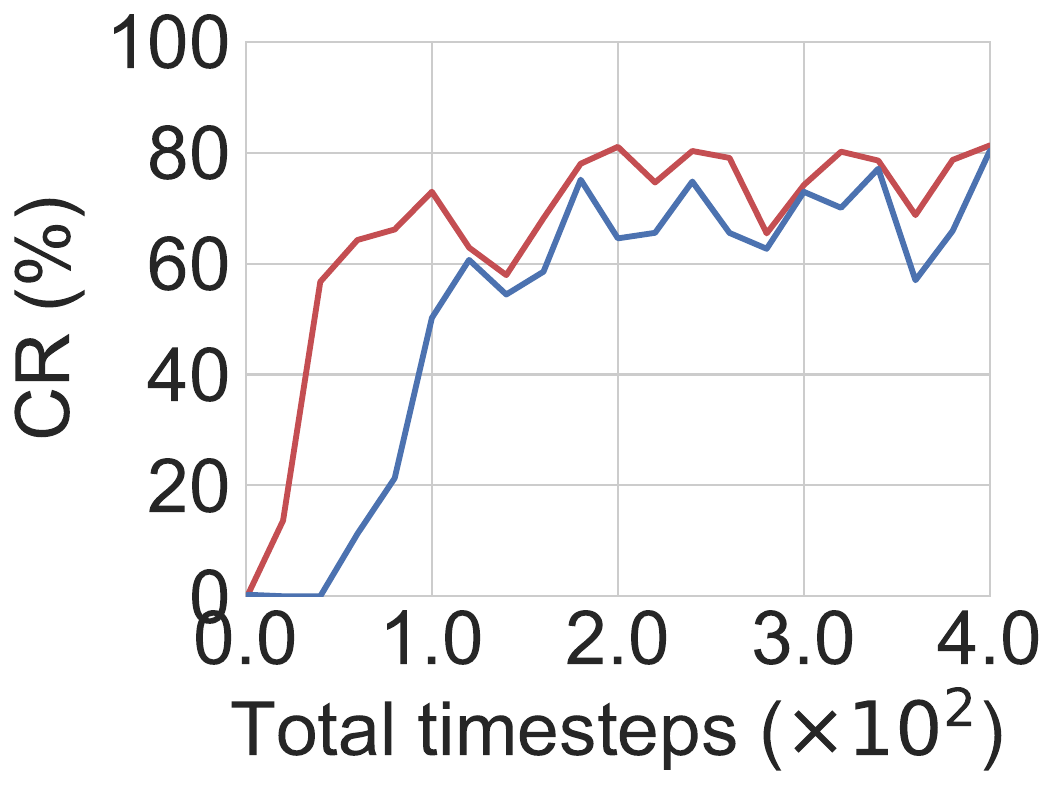}
    \caption{CR-322}
  \end{subfigure}
  \begin{subfigure}{0.22\textwidth}
    \centering
    \includegraphics[width=\textwidth]{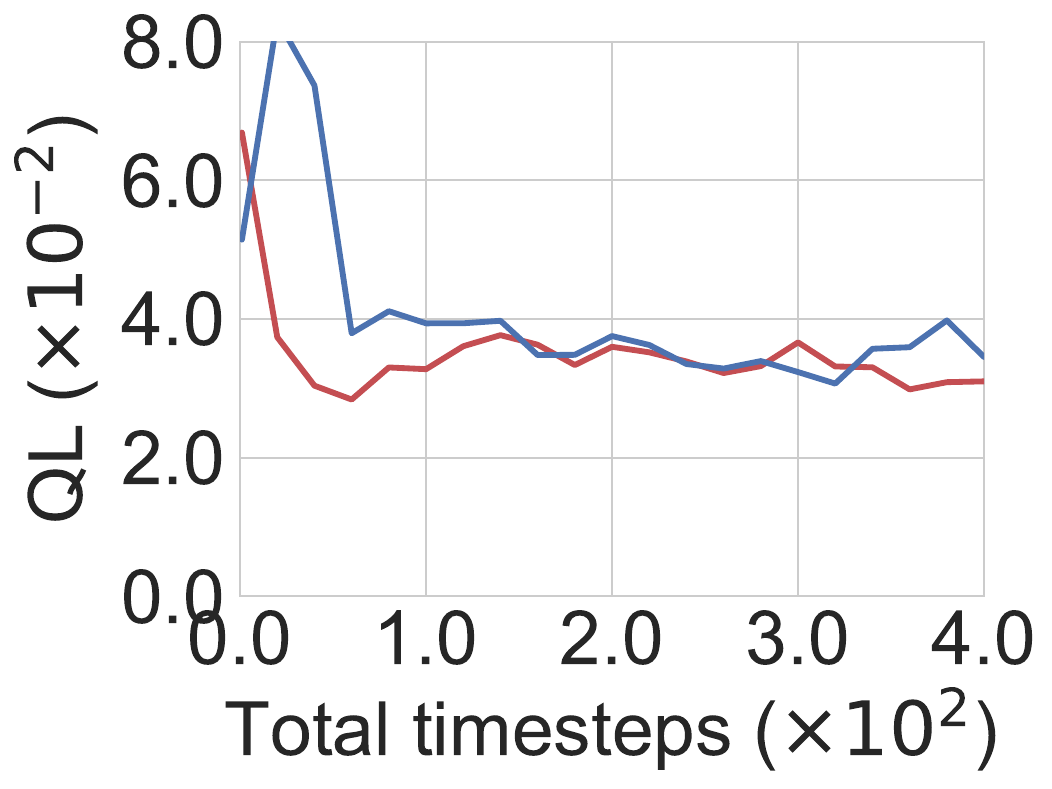} 
    \caption{QL-322}
  \end{subfigure}
  \caption{Training curves of TPA and randomly initialized prototypes on the IEEE 322-bus system.}
  \label{fig:randompp}
\end{figure}

\begin{table}[H]
\caption{Test results of TPA and randomly initialized prototypes on the 322-bus system. Bold denotes the best results. ± corresponds to the standard deviation of testing episodes.}
\resizebox{1.0\columnwidth}{!}{
  \begin{tabular}{@{}ll|cccccccccc@{}}
  \toprule
  \multicolumn{2}{c|}{\multirow{2}{*}{Methods}} & \multicolumn{2}{c}{Spring}             & \multicolumn{2}{c}{Summer}             & \multicolumn{2}{c}{Fall}               & \multicolumn{2}{c}{Winter}             & \multicolumn{2}{c}{All} \\ \cmidrule(l){3-12} 
  \multicolumn{2}{c|}{}                         & CR        & \multicolumn{1}{c|}{QL}    & CR        & \multicolumn{1}{c|}{QL}    & CR        & \multicolumn{1}{c|}{QL}    & CR        & \multicolumn{1}{c|}{QL}    & CR         & QL         \\ \midrule
  \multicolumn{2}{l|}{322-MADDPG}               & 80.1{\scalebox{0.8}{$\pm$18.9}} & \multicolumn{1}{c|}{0.033} & 74.6{\scalebox{0.8}{$\pm$19.8}} & \multicolumn{1}{c|}{0.033} & 92.7{\scalebox{0.8}{$\pm$14.0}} & \multicolumn{1}{c|}{0.032} & 96.7{\scalebox{0.8}{$\pm$7.6}}  & \multicolumn{1}{c|}{0.031} & 86.0         & 0.032      \\
  \multicolumn{2}{l|}{322-T-MADDPG}             & 87.8{\scalebox{0.8}{$\pm$19.8}} & \multicolumn{1}{c|}{0.032} & 83.3{\scalebox{0.8}{$\pm$20.7}} & \multicolumn{1}{c|}{0.033} & 96.2{\scalebox{0.8}{$\pm$8.7}}  & \multicolumn{1}{c|}{0.031} & 97.1{\scalebox{0.8}{$\pm$8.5}}  & \multicolumn{1}{c|}{0.030} & 91.1       & 0.031      \\ \midrule
  \multicolumn{2}{l|}{\cellcolor{gray!13}322-TPA-Random-MADDPG}        & 89.5{\scalebox{0.8}{$\pm$19.3}} & \multicolumn{1}{c|}{0.030} & 85.6{\scalebox{0.8}{$\pm$20.2}} & \multicolumn{1}{c|}{\textbf{0.030}} & 97.1{\scalebox{0.8}{$\pm$6.8}}  & \multicolumn{1}{c|}{\textbf{0.029}} & 96.4{\scalebox{0.8}{$\pm$10.3}} & \multicolumn{1}{c|}{\textbf{0.029}} & 92.1       & 0.030      \\
  \multicolumn{2}{l|}{\cellcolor{gray!13}322-TPA-MADDPG}           & \textbf{92.7}{\scalebox{0.8}{$\pm$17.1}} & \multicolumn{1}{c|}{\textbf{0.028}} & \textbf{91.1}{\scalebox{0.8}{$\pm$20.3}} & \multicolumn{1}{c|}{\textbf{0.030}} & \textbf{98.4}{\scalebox{0.8}{$\pm$5.7}}  & \multicolumn{1}{c|}{\textbf{0.029}} & \textbf{99.0}{\scalebox{0.8}{$\pm$3.3}}    & \multicolumn{1}{c|}{\textbf{0.029}} & \textbf{95.3}       & \textbf{0.029}      \\ \bottomrule
\end{tabular}
}
\label{tab:randompp}
\end{table}

In the original experiments, we directly adopted the load feature of buses to initialize the prototype. 
To study the adaptability and robustness of the prototypes, we have additionally conducted experiments to verify the performance of randomly initialized prototypes. As shown in Figure~\ref{fig:randompp} and Table~\ref{tab:randompp}. 
During the training phase, the initial fitting of the random prototype clearly requires more epochs.
Although the final performance of Random TPA slightly drops, it still obtains better results than baselines. PPA with random prototypes can still learn long-term temporal features without the strong requirements of manual design.
In future work, we can also adopt the concept of codebook learning to realize more effective learnable prototypes. We can apply factorized and normalized codes to ensure codebook quality with a larger prototype space.

\end{document}